\documentclass[manuscript,screen]{acmart}

\AtBeginDocument{%
  \providecommand\BibTeX{{%
    \normalfont B\kern-0.5em{\scshape i\kern-0.25em b}\kern-0.8em\TeX}}}

\usepackage{amsmath,amsfonts}
\usepackage{algorithmic}
\usepackage[linesnumbered,ruled,vlined]{algorithm2e}
\usepackage{amsthm}
\usepackage{graphicx}
\usepackage{textcomp}
\usepackage{paralist}
\usepackage{subcaption}
\usepackage{hyperref}
\usepackage{multirow}
\usepackage{url}
\usepackage{setspace}
\usepackage{tabu}
\usepackage{tablefootnote}
\usepackage{booktabs}
\usepackage{enumitem}
\usepackage{arydshln}
\usepackage[normalem]{ulem}
\usepackage{threeparttable}
\usepackage{flushend}
\usepackage{url}
\usepackage{xcolor}
\usepackage{soul}
\usepackage{natbib}

\newcommand{\tool}{SG-Trans\xspace}

\setcopyright{acmcopyright}
\acmJournal{TOSEM}
\acmYear{2022} \acmVolume{1} \acmNumber{1} \acmArticle{1} \acmMonth{1} \acmPrice{15.00}\acmDOI{10.1145/3522674}

\acmConference[Woodstock '18]{Woodstock '18: ACM Symposium on Neural
  Gaze Detection}{June 03--05, 2018}{Woodstock, NY}
\acmBooktitle{Woodstock '18: ACM Symposium on Neural Gaze Detection,
  June 03--05, 2018, Woodstock, NY}
\acmPrice{15.00}
\acmISBN{978-1-4503-XXXX-X/18/06}

\begin{document}

\title{Code Structure Guided Transformer for Source Code Summarization}

\author{Shuzheng Gao}
\email{szgao98@gmail.com}
\affiliation{%
  \institution{Harbin Institute of Technology, Shenzhen}
  \country{China}
}

\author{Cuiyun Gao}
\authornote{Corresponding author.}
\affiliation{%
  \institution{Harbin Institute of Technology, Shenzhen}
  \country{China}
  }
\email{gaocuiyun@hit.edu.cn}

\author{Yulan He}
\affiliation{%
  \institution{University of Warwick}
  \country{UK}
}
\email{yulan.he@warwick.ac.uk}

\author{Jichuan Zeng}
\affiliation{%
 \institution{The Chinese University of Hong Kong, Hong Kong}
  \country{China}}
  \email{jczeng@cse.cuhk.edu.hk}

\author{Lun Yiu Nie}
\affiliation{%
  \institution{Tsinghua University}
  \country{China}}
\email{nlx20@mails.tsinghua.edu.cn}

\author{Xin Xia}
\affiliation{%
  \institution{Software Engineering Application Technology Lab, Huawei}
  \country{China}}
\email{xin.xia@acm.org}

\author{Michael R. Lyu}
\affiliation{%
  \institution{The Chinese University of Hong Kong, Hong Kong}
  \country{China}}
\email{lyu@cse.cuhk.edu.hk}

\renewcommand{\shortauthors}{Gao, et al.}

\begin{abstract}
Code summaries
help developers comprehend programs and reduce their time to infer the program functionalities during software maintenance. 
Recent efforts resort to deep learning techniques such as sequence-to-sequence models for generating accurate code summaries, among which Transformer-based approaches have achieved promising performance. However, effectively integrating the code structure information into the Transformer is under-explored in this task domain.
In this paper, we propose a novel approach named \tool to incorporate code structural properties into Transformer. Specifically, we inject the local symbolic information (e.g., code tokens and statements) and global syntactic structure (e.g., data flow graph) into the self-attention module of Transformer as inductive bias. To further capture the hierarchical characteristics of code, the local information and global structure are designed to distribute in the attention heads of lower layers and high layers of Transformer.
Extensive evaluation shows the superior performance of \tool over the state-of-the-art approaches. Compared with the best-performing baseline, \tool still improves 1.4\% and 2.0\% in terms of METEOR score, a metric widely used for measuring generation quality, respectively on two benchmark datasets.

\end{abstract}




\maketitle

\section{Introduction}\label{sec:intro}
Program comprehension is crucial for developers during software development and maintenance. However, existing studies~\cite{DBLP:journals/tse/XiaBLXHL18, DBLP:conf/iwpc/MinelliML15} have shown that program comprehension is a very time-consuming activity which occupies over 50\% of the total time in software maintenance. To alleviate the developers' cognitive efforts in comprehending programs, a text summary accompanying the source code is proved to be useful~\cite{DBLP:conf/iwpc/HuLXLJ18,DBLP:journals/infsof/GarousiGRZMS15,DBLP:journals/jss/ChenH09}. However, 
human-written comment is often incomplete or outdated because of the huge effort it takes and the rapid update of software~\cite{DBLP:conf/sigdoc/SouzaAO05,DBLP:conf/fase/ShiZXL11}. 
The source code summarization task aims at automatically generating a concise comment of a program. 
Many studies~\cite{DBLP:conf/kbse/Wei19,DBLP:conf/acl/ChoiBNL21,DBLP:conf/iwpc/StapletonGLEWL020,DBLP:conf/iclr/LiuCXS021} have demonstrated that the machine-generated summaries are helpful for code comprehension.  A recent empirical study~\cite{DBLP:conf/icse/Huprac} also shows that 80\% of practitioners consider that code summarization tools can help them improve development efficiency and productivity.

\begin{figure}[t]
\centering
\includegraphics[width=0.7\textwidth]{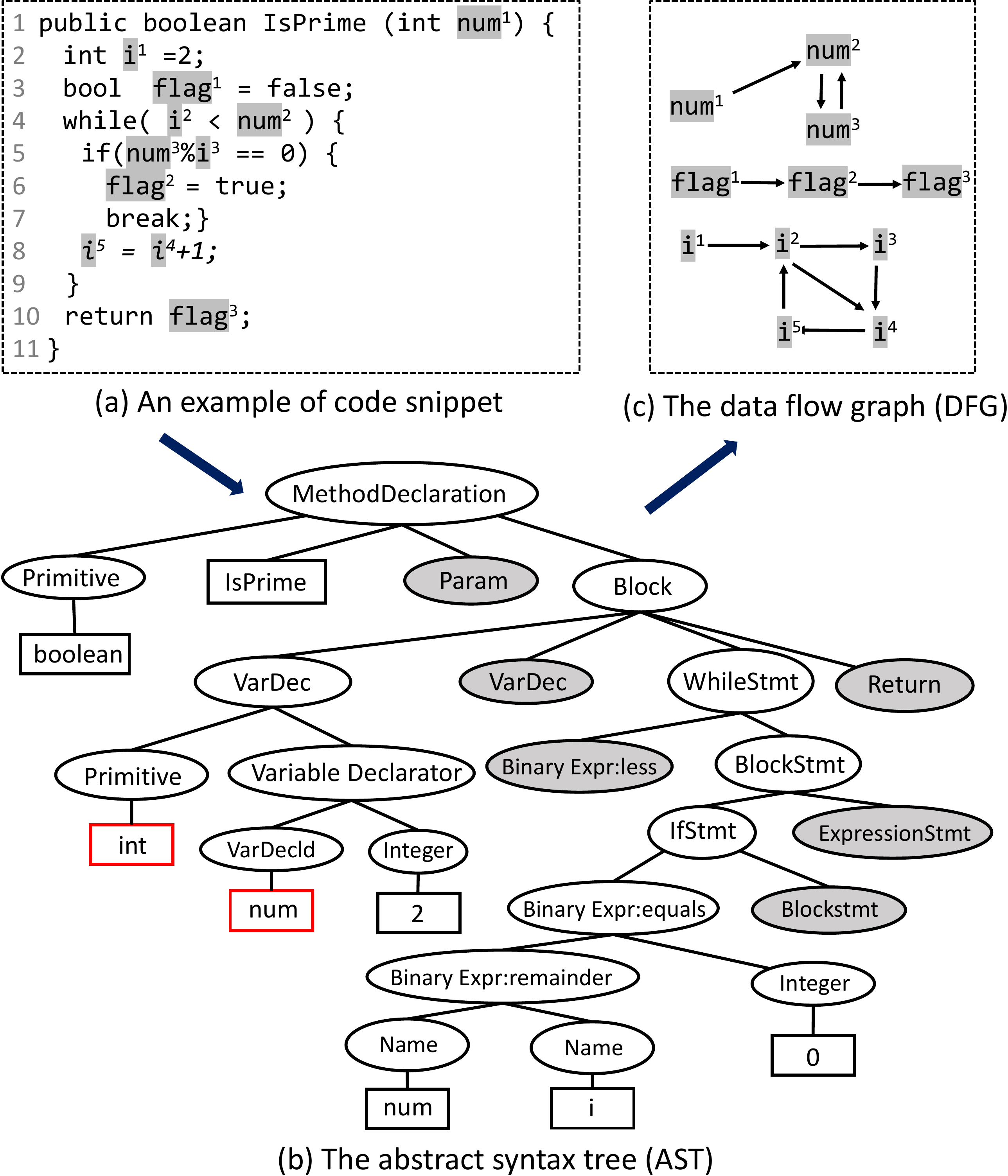}
\caption{An example of Java code snippet (a), with the corresponding AST (b) and DFG (c) illustrated. Entities in grey ellipse in (b) mean unexpanded branches. The arrows in the DFG represent the relations of sending/receiving messages between the variables (highlighted in grey in the code).}
\label{fig:data_flow}
\end{figure}

Existing leading approaches have demonstrated the benefits of integrating code structural properties such as Abstract Syntax Trees (ASTs)~\cite{DBLP:conf/iclr/AlonBLY19,DBLP:conf/iwpc/HuLXLJ18} into deep learning techniques for the task. An example of AST is shown in Figure~\ref{fig:data_flow} (b). The modality of the code structure can be either sequences of tokens traversed from the syntactic structure of ASTs~\cite{DBLP:conf/iclr/AlonBLY19,DBLP:conf/iwpc/HuLXLJ18} or sequences of small statement trees split from large ASTs  ~\cite{DBLP:conf/icse/ZhangWZ0WL19,DBLP:conf/ijcnn/ShidoKYMM19}. The sequences are usually fed into a Recurrent Neural Network (RNN)-based sequence-to-sequence network for generating a natural language summary~\cite{DBLP:conf/iwpc/HuLXLJ18,DBLP:conf/aaai/LiangZ18}. However, due to the deep nature of ASTs, the associated RNN-based models may fail to capture the long-range dependencies between code tokens~\cite{DBLP:conf/acl/AhmadCRC20}. To mitigate this issue, some works represent the code structure as graphs and adopt Graph Neural Networks (GNNs) for summary generation~\cite{DBLP:conf/iclr/FernandesAB19,DBLP:conf/iwpc/LeClairHWM20}. Although these GNN-based approaches can capture the long-range relations between code tokens, they are shown 
sensitive to local information and ineffective in capturing the global structure~\cite{DBLP:journals/ijgi/IddianozieM20}. Taking the AST in Figure~\ref{fig:data_flow} (b) as an example, token nodes ``\textit{int}'' and ``\textit{num}'' (highlighted with red boxes) are in the same statement but separated by five hops, so GNN-based approaches tend to ignore the relations between the two token nodes. Besides, the message passing on GNNs is limited by the pre-defined graph, reducing its scalability to learn other dependency relations.

Recent study~\cite{DBLP:conf/acl/AhmadCRC20} shows that the Transformer model~\cite{DBLP:conf/nips/VaswaniSPUJGKP17} outperforms other deep learning approaches for the task. The self-attention mechanism in Transformer can be viewed as a fully-connected graph~\cite{DBLP:conf/naacl/GuoQLSXZ19}, which can ensure the long-range massage passing between tokens and the flexibility to learn any dependency relation from data. However, it is hard for the Transformer to learn all important dependency relations from limited training data. Besides, an issue of Transformer is that its attention is purely data-driven~\cite{DBLP:conf/aaai/GuoQLXZ20}. Without {the incorporation of} explicit constraints, the multi-head attentions in Transformer may suffer from attention collapse or attention redundancy, {with} different attention heads extracting similar attention features, {which hinders} the model's representation {learning ability}~\cite{DBLP:conf/acl/VoitaTMST19,DBLP:conf/emnlp/AnLWLHTZHC20}. To solve the aforementioned problems, we incorporate code structure into the Transformer as prior information to eliminate its dependency on data. However, how to effectively integrate the code structure information into Transformer is still under-explored. One major challenge is that since the position encoding in the Transformer already learns the dependency relations between code tokens, trivial integration of the structure information may not bring an improvement for the task~\cite{DBLP:conf/acl/AhmadCRC20}.

To overcome the above challenges
in this paper, we propose a novel model named \tool, i.e., code \textbf{S}tructure \textbf{G}uided \textbf{Trans}former. \tool exploits the code structural properties to introduce explicit constraints to the multi-head self-attention module. Specifically, we extract the pairwise relations between code tokens based on the local symbolic structure such as code tokens and statements, and the global syntactic structure, i.e., data flow graph (DFG), then represent them as adjacency matrices before injecting them into the multi-head attention mechanism as inductive bias. Furthermore, following the principle of compositionality in language: the high-level semantics is the composition of low-level terms~\citep{DBLP:conf/aaai/GuoQLXZ20,sep-compositionality}, we propose a hierarchical structure-variant attention approach to guide the attention heads at the lower layers attending more to the local structure and {those at the} higher layers attending more to the global structure. In this way, our model can take advantage of both local and global (long-range dependencies) information of source code. Experiments on benchmark datasets demonstrate that \tool can outperform the state-of-the-art models by at least 1.4\% and 2.0\% in terms of METEOR on two Java and Python benchmark datasets, respectively.

{In summary,} our work makes the following contributions:

\begin{itemize}
  \item We are the first to explore the integration of both local and global code structural properties into Transformer for source code summarization.
  \item A novel model is proposed to hierarchically incorporate both the local and global structure of code into the multi-head attentions in Transformer as inductive bias.
  \item Extensive experiments show \tool outperforms the state-of-the-art models. We publicly release the replication repository including source code, datasets, prediction logs, online questionnaire, and results of human evaluation on GitHub\footnote{https://github.com/gszsectan/SG-Trans}.
\end{itemize}

\textbf{Paper structure.} Section~\ref{sec:background} illustrates the background knowledge of the work.
Section~\ref{sec:approach} presents 
{our proposed} methodology 
for source code summarization. Section~\ref{sec:setup} introduces the experimental setup. Section~\ref{sec:results} describes the evaluation results, followed by {the discussions in} Section~\ref{sec:discussion}. 
Section~\ref{sec:literature} presents related studies. 
{Finally,} Section~\ref{sec:con} {concludes the paper and outlines future research work}.

 \section{Background}\label{sec:background}
 
In this section, we introduce the background knowledge of the proposed approach, including the vanilla Transformer model architecture and the copy mechanism.

\subsection{Vanilla Transformer}

Transformer~\cite{DBLP:conf/nips/VaswaniSPUJGKP17} is a kind of deep self-attention network which has demonstrated its powerful text representation capability in many NLP applications, e.g., machine translation and dialogue generation~\cite{DBLP:conf/aaai/SongWY0HLDZ20,DBLP:conf/www/ZhaoWHYCW20}. Recently, a lot of research in code summarization also leverage Transformer as backbone for better source code representations~\cite{DBLP:conf/acl/AhmadCRC20,DBLP:conf/emnlp/FengGTDFGS0LJZ20}. Some work \cite{DBLP:conf/iclr/HellendoornSSMB20,DBLP:conf/iclr/ZugnerKCLG21,DBLP:conf/acl/WangSLPR20} also improves the Transformer to make it  better adapt to source code or structured data. Unlike conventional neural networks such as Convolutional Neural Network (CNN) and Recurrent Neural Network (RNN), it is solely based on attention mechanism and multi-layer perceptrons (MLPs). Transformer follows the sequence-to-sequence~\cite{DBLP:conf/emnlp/ChoMGBBSB14} architecture with stacked encoders and decoders. Each encoder block and decoder block consist of a multi-head self-attention sub-layer and a feed-forward sub-layer. Residual connection~\cite{DBLP:conf/cvpr/HeZRS16} and layer normalization~\cite{DBLP:journals/corr/BaKH16} are also employed between the sub-layers. Since the two sub-layers play an essential role in Transformer, We introduce them in more detail as the following.

\subsubsection{Multi-Head Self-Attention} Multi-head attention is the key component of Transformer. Given an input sequence $X=(x_1,x_2,...,x_i,...,x_n)$ where $n$ is the sequence length and each input token $x_i$ is represented by a $d$-dimension vector, self-attention first calculates the Query vector, the Key vector, and the Value vector for each input token by multiplying the input vector with three matrices $W^q$, $W^k$, $W^v$. Then it calculates the attention weight of sequence $X$ by scoring the query vector $Q$ against the key vector $K$ of the input sentence. The scoring process is conducted by the scaled dot product, as shown in Eq. (\ref{equ:self_atten}), where the dimension $d$ in the denominator is used to scale the dot product. Softmax is then used to normalize the attention score and finally the output vectors is computed as a weighted sum of the input vectors.
Instead of performing a single self-attention function, Transformer adopts multi-head self-attention (MHSA) which performs the self-attention function with different parameters in parallel and ensembles the output of each head by concatenating their outputs. The MHSA allows the model to jointly attend to information from different representation subspaces at different positions. Formally, the MHSA is computed as following:
\begin{equation}
    Q_i = XW_i^q, \quad K_i = XW_i^k, \quad V_i = XW_i^v,
\end{equation}
\begin{equation}\label{equ:self_atten}
    head_i = \operatorname{softmax}(\frac{Q_iK_i^T}{\sqrt{d}})V_i,
\end{equation}
\begin{equation}
    MHSA(X) = [head_1^l\circ head_2^l \circ\cdots head_i^l \circ \cdots head_h^l]W^O,
\end{equation}
\noindent where $h$ denotes the number of attention heads at $l$-th each layer, the symbol $\circ$ indicates the concatenation of $h$ different heads, and $W_i^q$, $W_i^k$, $W_i^v$ and $W^O$ are trainable parameters.

\subsubsection{Feed-Forward Network} Feed-forward network is the only nonlinear part in Transformer. It consists of two linear transformation layer and a ReLU activation function between the two linear layers:
\begin{equation}
    FFN(X) = ReLU(XW_1+b_1)W_2+b_2,
\end{equation}
\noindent where $W_1$, $W_2$, $b_1$, and $b_2$ are trainable parameters which are shared across input positions.
\\

\begin{figure}[t]
\centering
\includegraphics[width=0.8\textwidth]{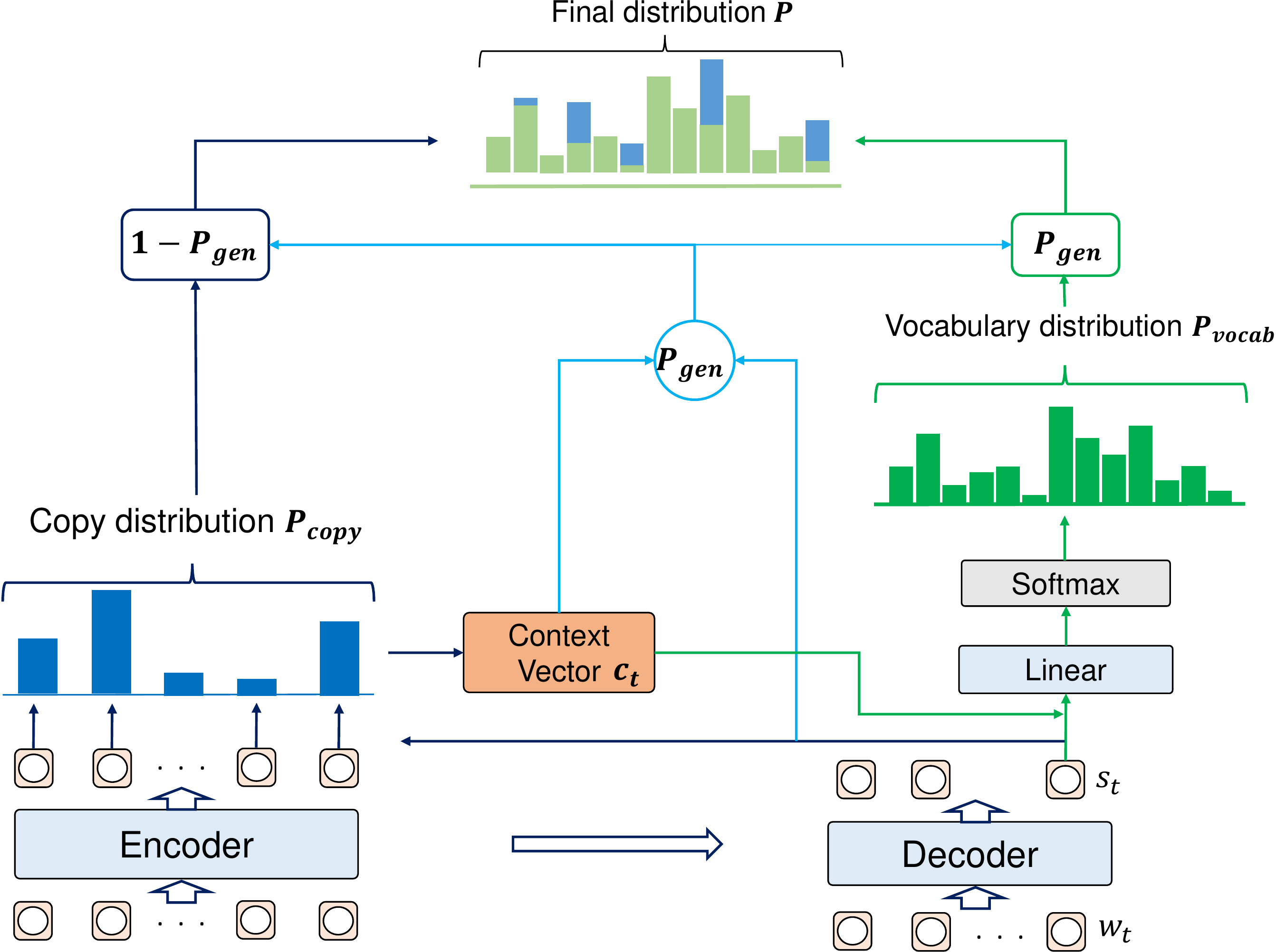}
\caption{Architecture of copy mechanism.}
\label{fig:copy}
\end{figure}

\subsection{Copy Mechanism}\label{subsec:copy}

Copy mechanism~\cite{DBLP:conf/acl/GuLLL16} has been widely equipped in text generation models for extracting words from a source sequence as part of outputs in a target sequence during text generation. It has been demonstrated that copy mechanism can alleviate the out-of-vocabulary issue in the code summarization task~\cite{DBLP:conf/iclr/ZugnerKCLG21,DBLP:conf/acl/AhmadCRC20,DBLP:conf/icse/ZhangW00020}. Besides, copying some variable name can also help generate more precise summary. In this work, we adopt the pointer generator~\cite{DBLP:conf/acl/SeeLM17}, a more popular form of copy mechanism, for the task. Figure~\ref{fig:copy} illustrates the architecture of the pointer generator model. Specifically, given an input sequence $X=(x_1,x_2,...,x_n)$, a decoder input $w_t$, a decoder hidden state $s_t$, and a context vector $c_t$ computed by the attention mechanism in time step $t$, the pointer generator first calculates a constant $P_{gen}$ which is later used as a soft switch for determining whether to generate a token from the vocabulary or to copy a token from the input sequence ${X}$:

\begin{equation}
    P_{gen} = \operatorname{sigmoid}(\omega_s^\mathsf{T} s_t+\omega_w^\mathsf{T} w_{t}+\omega_c^\mathsf{T} c_t+b_{gen}),
\end{equation}
\begin{equation}
    P_{vocab}(w_t) = \operatorname{softmax}(W_as_t+V_ac_t)
\end{equation}
\begin{equation}
    P(w_t)=P_{gen}P_{vocab}(w_t)+(1-P_{gen})P_{copy}(w_t ),
\end{equation}

\noindent where vectors $\omega_s$, $\omega_w$, $\omega_c$, $W_a$, $V_a$ and scalar $b_{gen}$ are learnable parameters. $P(w_t)$ is the probability distribution over the entire vocabulary. Copy distribution $P_{copy}(w_t)$ determines where to attend to in time step $t$, computed as:

\begin{equation}
    P_{copy}(w_t) = \sum_{i:x_i=w} \alpha_{t,i}\ ,
\end{equation}

\noindent where $\alpha_{t}$ indicates the attention weights and $i:x_i=w$ indicates the indices of input words in the vocabulary.

\section{Proposed Approach}\label{sec:approach}

In this section, we explicate the detailed
architecture of \tool. Let $D$ denotes a dataset containing a set of programs $C$ and their associated summaries $Z$, given source code $c=(x_1, x_2, ..., x_n)$ from $C$, where $n$ denotes the code sequence length. \tool is designed to generate the summary consisting of a sequence of tokens $\hat{z}=(y_1, y_2, ..., y_m)$ by maximizing the conditional likelihood: $\hat{z}=\mathop{\arg\max}_zP(z|c)$ ($z$ is the corresponding summary in $Z$).


The framework of \tool is mostly consistent with the vanilla Transformer, but consists of two major improvements, namely \textit{structure-guided self-attention} and \textit{hierarchical structure-variant attention}. Figure \ref{fig:framework} depicts the overall architecture. \tool first parses the input source code for capturing both local and global structure. The structure information is then represented as adjacency matrices and incorporated into the self-attention mechanism as inductive biases (introduced in Section~\ref{subsec:struct_guided}). Following the principle of compositionality in language, 
different inductive biases are 
integrated into the Transformer at difference levels in a hierarchical manner
(introduced in Section~\ref{subsec:hierarch}).


\subsection{Structure-Guided Self-Attention}\label{subsec:struct_guided}

\begin{figure*}[t]
\centering
\includegraphics[width=0.97\textwidth]{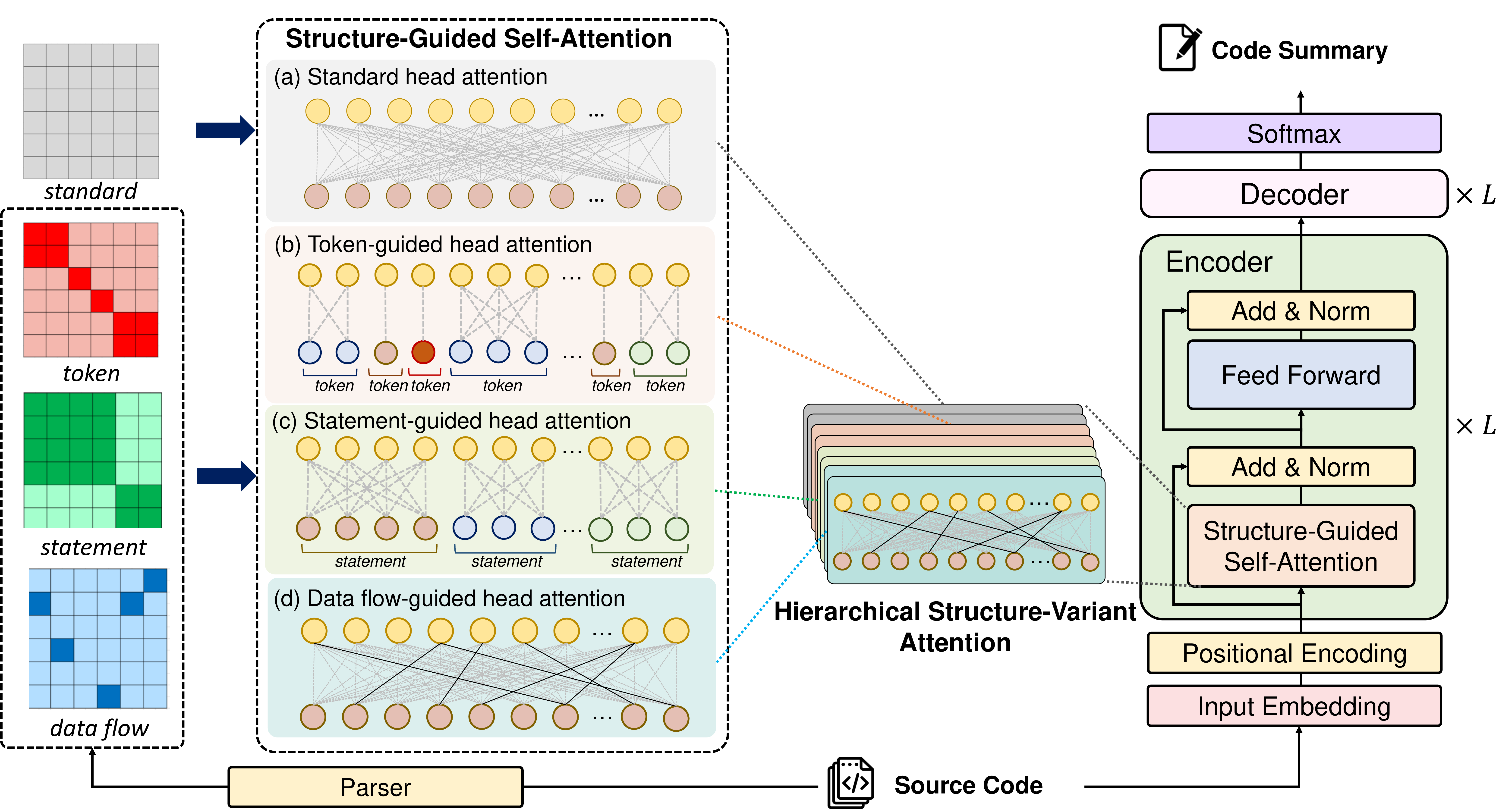}
\caption{Overall framework of the proposed \tool. The ``Structure-Guide Self-Attention'' part illustrates different self-attention mechanisms between adjacent layers.}
\label{fig:framework}
\end{figure*}

In the standard multi-head self-attention model
\citep{DBLP:conf/nips/VaswaniSPUJGKP17}, 
every node in the adjacent layer is allowed to attend to all the input nodes, as shown in Figure~\ref{fig:framework} (a). 
In this work, we propose to use the {structural} relations in source code to introduce explicit constraints to the multi-head self-attention. In order to capture the hierarchical structure of source code, we utilize three main types of structural relations between code tokens, including local structures: whether the {two split sub-tokens originally} belong to the same 1) \textbf{token} or 2) \textbf{statement}, and global structure: whether there exists a 3) \textbf{data flow} between two tokens.
For each structure type, we design the corresponding head attention, named as \textit{token-guided self-attention}, \textit{statement-guided self-attention}, and \textit{data flow-guided self-attention}, respectively.

\textbf{Token-guided self-attention.} 
Semantic relations between the sub-tokens are relatively stronger than the relations between the other tokens. For example, the method name ``\textit{IsPrime}'' in the code example shown in Figure~\ref{fig:data_flow} (a) is split as a sequence of sub-tokens containing ``\textit{Is}'' and ``\textit{Prime}''. Moreover, the semantic relation between them is stronger than the relation between ``\textit{Is}'' and ``\textit{num}'' in the same statement.
Therefore, the attention can be built upon the extracted token-level structure, i.e, whether two 
sub-tokens are originally from the same source code token. We use an adjacency matrix $\mathbf{T}^{n\times n}$ to model the relationship, where $t_{ij}=0$ if the $i$-th and $j$-th elements are sub-tokens of the same token in the code; Otherwise, $t_{ij}=-\infty$. The matrix is designed to restrict the attention head to only attend to the sub-tokens belonging to the same code token in self-attention, as shown in Figure~\ref{fig:framework} (b). Given the input token representation $\mathbf{X}\in \mathbb{R}^{n\times dh}$, where $n$ is the sequence length, $d$ is the input dimension of each head, and $h$ is the number of attention heads. Let $\mathbf{Q}$, $\mathbf{K}$, and $\mathbf{V}$ denote the query, key, and value matrix, respectively, the token-guided single-head self-attention $head_t$ can be calculated as:


\begin{equation}
{ head }_{t}=\operatorname{softmax}\left(\frac{\mathbf{Q} \mathbf{K}^\intercal}{\sqrt{d}}+\mathbf{T}\right)\mathbf{V},
\end{equation}

\noindent where 
$\sqrt{d}$ is a scaling factor to prevent the effect of large values.

\textbf{Statement-guided self-attention.} Tokens in the same statement tend to possess stronger semantic relations than those from different statements. For the code example given in Figure~\ref{fig:data_flow} (a), the token ``\textit{flag}'' in the third statement is more relevant to the tokens ``\textit{bool}'' and ``\textit{False}'' in the same statement than to the token ``\textit{break}'' in the 7-th statement. So we design another adjacency matrix $S$ to represent the pairwise token relations capturing whether the two tokens are from the same statement. In the matrix $\mathbf{S}$, $s_{ij}=0$ if the $i$-th and $j$-th input tokens are in the same statement; otherwise, $s_{ij}=-\infty$. The design is to restrict the attention head to only attend to the tokens from the same statement, as illustrated in Figure~\ref{fig:framework} (c). The statement-guided single-head self-attention $head_s$ is defined below, similar to the token-guided head attention:


\begin{equation}
{ head }_{s}=\operatorname{softmax}\left(\frac{\mathbf{Q} \mathbf{K}^\intercal}{\sqrt{d}}+\mathbf{S}\right)\mathbf{V}.
\end{equation}

\textbf{Data flow-guided self-attention.} To facilitate the model to
learn the global semantic information from code, we employ the data flow graphs (DFGs) for capturing the global semantic structure feature~\cite{DBLP:journals/corr/abs-2009-08366}. We do not involve ASTs as input since they are deeper in nature and contain more redundant information \cite{DBLP:conf/iclr/AllamanisBK18} than DFGs. DFGs, denoted as $V=\{v_1, v_2,...\}$, can model the data dependencies between variables in the code, including message sending/receiving. Figure~\ref{fig:data_flow} (c) shows an example of the extracted data flow graph. Variables with same name (e.g., \textit{i}\textsuperscript{2} and \textit{i}\textsuperscript{5}) are associated with different semantics in the DFG. Each variable is a node in the DFG and the direct edge $\langle v_i, v_j \rangle$ from $v_i$ to $v_j$ indicates the value of the $j$-th variable comes from the $i$-th variable. We can find that the semantic relations among ``\textit{i}\textsuperscript{2}'', ``\textit{i}\textsuperscript{3}'', ``\textit{i}\textsuperscript{4}'' and ``\textit{i}\textsuperscript{5}'' which represent the data sending/receiving in a loop.
Based on the DFGs, we build the adjacency matrix $\mathbf{D}$, where $d_{ij}=1$ if there exists a message passing from the $j$-th token to the $i$-th token; Otherwise, $d_{ij}=0$. Note that if two variables have a data dependency, then their constituent sub-tokens also possess the dependency relation. Figure~\ref{fig:framework} (d) illustrates the data flow-guided single-head self-attention. To address the sparseness of the matrix $\mathbf{D}$ and to highlight the relations of data dependencies, we propose the data flow-guided self-attention $head_f$ below:

\begin{equation}\label{equ:dataflow}
    head_f = \operatorname{softmax}\left(\frac{\mathbf{Q} \mathbf{K}^\intercal+\mu*\mathbf{Q} \mathbf{K}^\intercal\mathbf{D}}{\sqrt{d}}\right)\mathbf{V},
\end{equation}

\noindent where $\mu$ is the control factor for adjusting the integration degree of the data flow structure. 

\subsection{Hierarchical Structure-Variant Attention}\label{subsec:hierarch}

\begin{figure}[t]
\centering
\includegraphics[width=0.5\textwidth]{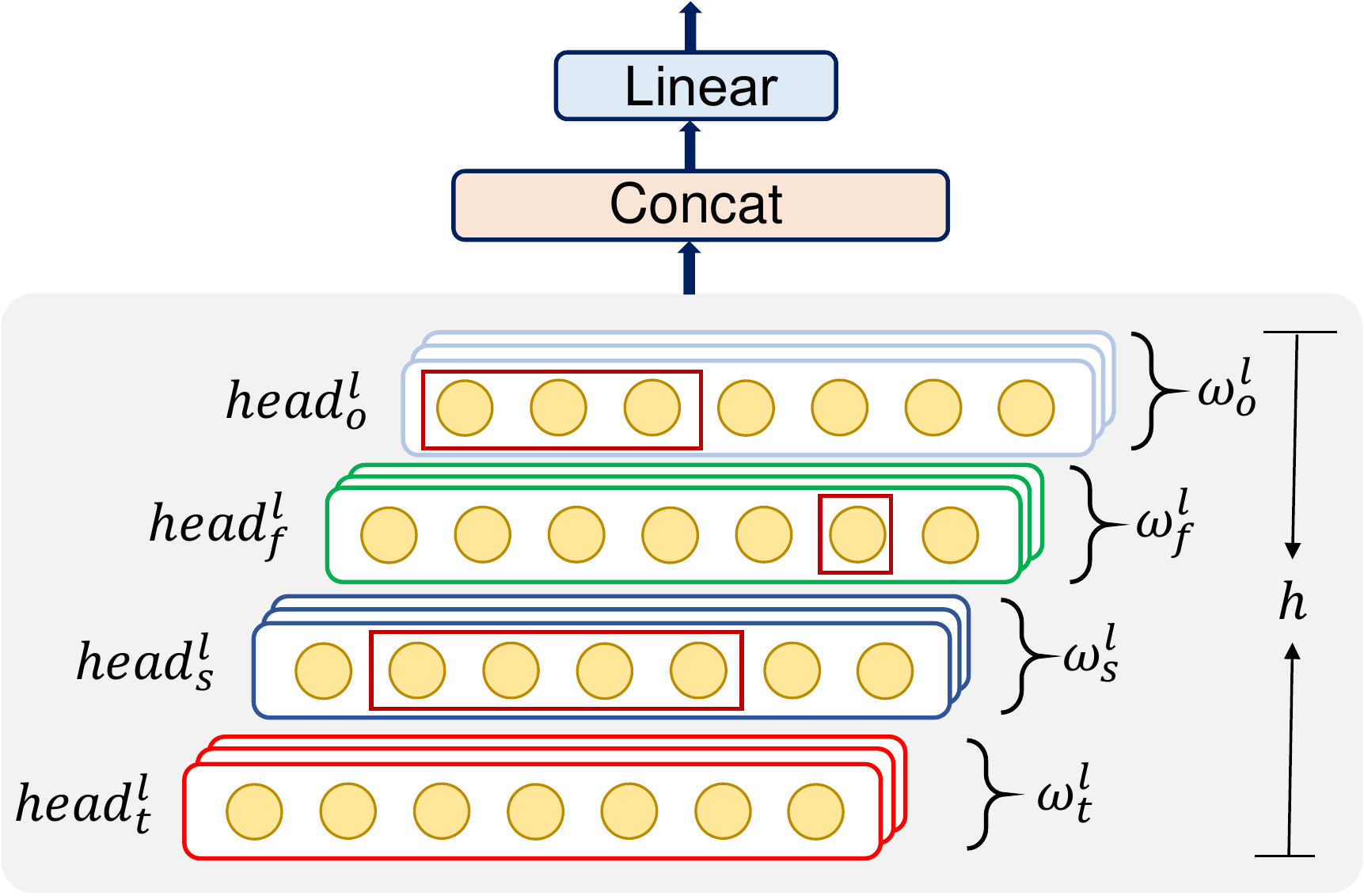}
\caption{A diagram of hierarchical-variant attention. Different red boxes illustrate different scales.}
\label{fig:multi_head}
\end{figure}

Inspired by the principle of compositionality in logic semantics: the high-level semantics is the composition of low-level terms \citep{DBLP:conf/aaai/GuoQLXZ20,sep-compositionality}, we propose a hierarchical structure-variant attention such that our model would focus on local structures at the lower layers and global structure at the higher layers. The diagram of the hierarchical structure-variant attention is illustrated in Figure~\ref{fig:multi_head}. Specifically, the token-guided head attention $head_t$ and the statement-guided head attention $head_s$ are used more in the heads of lower layers; while the data flow-guided head attention $head_f$ is more spread in the heads of higher layers.

Let $L$ denote the number of layers in the proposed \tool, $h$ indicate the number of heads in each layer and $k$ be a hyper-parameter to control the distribution of four types of head attentions, including $head_t$, $head_s$, $head_f$, and $head_o$, where $head_o$ indicates the standard head attention without {constraints}, 
the distribution for each type of head attention at the $l$-th layer is denoted as $\Omega^l=[\omega_t^l, \omega_s^l, \omega_f^l, \omega_o^l]$, where $\omega_t^l$, $\omega_s^l$, $\omega_f^l$, and $\omega_o^l$ represent the numbers of $head_t$, $head_s$, $head_f$, and $head_o$, respectively at the $l$-th layer. We define the distribution below:

\begin{equation}
    \omega_t^l = \omega_s^l = \lfloor h* \frac{k-l}{2*k-l}\rfloor,
\end{equation}
\begin{equation}
    \omega_f^l = \lfloor h* \frac{l}{2*k-l}\rfloor,
\end{equation}
\begin{equation}
    \omega_o^l = h-(\omega_t^l+\omega_s^l+\omega_f^l),
\end{equation}

\noindent where $k$ is a positive integer hyperparameter, and
$\lfloor \cdot \rfloor$ denotes rounding the value down to the next lowest integer. The design is to {enable} 
more heads 
attending to the global structure with the growth of $l$, i.e., $\omega^l_f$ will get larger at a higher layer $l$; meanwhile few heads can catch the local structure, i.e., $\omega^l_t$ and $\omega_s^l$ will become smaller. $head_o$ is involved to enable the model to be adapted to arbitrary numbers of layers and heads. Especially, with the increase of layer $l$, $\omega^l_t$ and $\omega^l_s$ might drop to zero. In the case of $\omega^l_t$  $\leq$ 0, no {constraints} 
will be introduced to the corresponding attention layer since the standard self-attention already captures long-range dependency information, which fits our purpose of attending to global structure at higher layers. Otherwise, the head attentions will follow the defined distribution $\Omega^l$.

The hierarchical structure-variant attention (HSVA) at the $l$-th layer is computed as:
\begin{equation}
    \text{HSVA}^l = [head_1^l\circ\cdots\circ head_h^l]\mathbf{W}^O,
\end{equation}

\noindent where $\circ$ denotes the concatenation of $h$ different heads, and $\mathbf{W}^O\in\mathbb{R}^{dh\times dh}$ is a parameter matrix.

\subsection{Copy Attention}
The OOV issue is important for effective code summarization~\citep{DBLP:conf/icse/KarampatsisBRSJ20}. We adopt the copy mechanism introduced in Section~\ref{subsec:copy} in \tool to calculate whether to generate words from {the} vocabulary or to copy from the input source code. Following Ahmad et al.~\citep{DBLP:conf/acl/AhmadCRC20}, an additional attention layer is added to learn the copy distribution on top of the decoder~\citep{DBLP:conf/acl/NishidaSNSOAT19}. The mechanism enables the proposed \tool to copy low-frequency words, e.g., API names, from source code, thus mitigating {the} OOV issue.

\section{Experiemental Setup}\label{sec:setup}
In this section, we introduce the evaluation datasets and metrics, comparison baselines, and parameter settings.

\begin{table}[t]
 \centering
 \aboverulesep=0ex
\belowrulesep=0ex
\caption{Statistics of the benchmark datasets.}
\label{tab:dataset}
\scalebox{1.0}{
\begin{tabular}{l|r|r}
\toprule
 & \textbf{ Java }   & \textbf{ Python } \\ 
\midrule
{\ Training Set} & 69,708 & 55,538 \\
{\ Validation Set } & 8,714 & 18,505 \\
{\ Test Set } & 8,714 & 18,502 \\ 
\midrule
{\ Total} & 87,136 & 92,545 \\ 
\bottomrule
\end{tabular}}
\end{table}

\subsection{Benchmark Datasets}
We conduct experiments on two benchmark datasets that respectively contain Java and Python source code following the previous work~\citep{DBLP:conf/acl/AhmadCRC20,DBLP:conf/icse/ZhangW00020}. 
Specifically, the Java dataset publicly released by Hu et al. \citep{DBLP:conf/iwpc/HuLXLJ18} comprises $87,136$ $\langle$Java method, comment$\rangle$ pairs collected from $9,714$ GitHub repositories, and the Python dataset consists of $92,545$ functions and corresponding documentation as originally collected by Barone et al. \citep{barone2017parallel} and later processed by Wei et al. \citep{DBLP:conf/kbse/Wei19}. 

For fair comparison, we directly use the benchmarks open sourced by the previous related studies \citep{DBLP:conf/ijcai/HuLXLLJ18,DBLP:conf/acl/AhmadCRC20,DBLP:conf/nips/WeiL0FJ19}, in which the datasets are split into training set, validation set, and test set in a proportion of $8:1:1$ and $6:2:2$ for Java and Python, respectively. We follow the commonly-used dataset split strategy with no modification to avoid any bias introduced by dataset split.

We apply \textit{CamelCase} and \textit{snake\_case} tokenizers~\citep{DBLP:conf/acl/AhmadCRC20} to get sub-tokens for both dataset. As for the code statements, we apply a simple rule to extract the statements of code snippet. For the extraction of statements from Java dataset, we split each
code snippet into statements with separators including '\{', '\}' and ';'. The token sequence between two adjacent separators is considered as a statement. For the example shown in Figure~\ref{fig:data_flow} (a), each line except the separators is one statement. While for the Python dataset, we define a statement by the row, which means that the tokens in the same row are considered as belonging to the same statement. For the extraction of data flow from the Java dataset, we use the tool in GSC \citep{DBLP:conf/icml/CvitkovicSA19} to first generate augmented ASTs and then from which extract DFGs. Regarding the Python dataset, we follow the setup in Allamanis et al.'s work~\citep{DBLP:conf/iclr/AllamanisBK18} and extract four kinds of edge (\textit{LastRead}, \textit{LastWrite}, \textit{LastLexicalUse}, \textit{ComputeFrom}) from code.


\subsection{Evaluation Metrics}
To verify the superiority of \tool over the baselines, we use the most commonly-used automatic evaluation metrics, BLEU-4 \citep{DBLP:conf/acl/PapineniRWZ02}, METEOR \citep{DBLP:conf/acl/BanerjeeL05} and ROUGE-L \citep{lin-2004-rouge}.

\textbf{BLEU} is a metric widely used in natural language processing and software engineering fields to evaluate generative tasks (e.g., dialogue generation, code commit message generation, and pull request description generation)~\citep{DBLP:conf/naacl/ZhangUSNN18,DBLP:conf/kbse/JiangAM17,DBLP:conf/kbse/Liu0T0L19,DBLP:journals/corr/abs-2007-06934}. BLEU uses $n$-gram for matching and calculates the ratio of $N$ groups of word similarity between generated comments and reference comments. The score is computed as:
\begin{equation}
    BLEU-N = BP \times \exp(\sum_{n=1}^{N}\tau_n \log P_n),
\end{equation}

\noindent where $P_n$ is the ratio of the subsequences with length $n$ in the candidate that are also in the reference. $BP$ is the brevity penalty for short generated sequence and $\tau_n$ is the uniform weight $1/N$. We use corpus-level BLEU-4, i.e., $N=4$, as our evaluation metric since it is demonstrated to be more correlated with human judgements than other evaluation metrics~\citep{DBLP:conf/emnlp/LiuLSNCP16}. 


\textbf{METEOR}
 is a recall-oriented metric which measures how well our model captures content from the reference text in our generated text. It evaluates generated text by aligning them to reference text and calculating sentence-level similarity scores.
\begin{equation}
    METEOR = (1-\gamma \cdot \textit{frag}^\beta) \cdot \frac{P \cdot R}{\alpha \cdot P + (1-\alpha) \cdot R},
\end{equation}
where P and R are the unigram precision and recall, \textit{frag} is the
fragmentation fraction. $\alpha$, $\beta$ and $\gamma$ are three penalty parameters whose default values are 0.9, 3.0 and 0.5, respectively.

\textbf{ROUGE-L} is widely used in text summarization tasks in the natural language processing field to evaluate what extent the reference text is recovered or captured by the generated text. ROUGE-L is based on the Longest Common Subsequence (LCS) between two text and the F-measure is used as its value. Given a generated text $X$ and the reference text $Y$ whose lengths are $m$ and $n$ respectively, ROUGE-L is computed as:
\begin{equation}
    P_{lcs} = \frac{LCS(X,Y)}{n}, R_{lcs} = \frac{LCS(X,Y)}{m},
    F_{lcs} = \frac{(1+\beta ^2)P_{lcs}R_{lcs}}{R_{lcs}+\beta^2 P_{lcs}},
\end{equation}
where $\beta = P_{lcs}/R_{lcs}$ and $F_{lcs}$ is the computed ROUGE-L value.

\subsection{Baselines}
We compare \tool with following baseline approaches.

{\textbf{\underline{CODE-NN}}} \cite{DBLP:conf/acl/IyerKCZ16}, as the first deep-learning-based work in code summarization, generates source code summaries with an LSTM network. To utilize code structure information, \textbf{\underline{Tree2seq}} \cite{DBLP:conf/acl/EriguchiHT16} encodes source code with a tree-LSTM architecture. \textbf{\underline{Code2seq}~\cite{DBLP:conf/iclr/AlonBLY19}} represents the code snippets by sampling paths from the AST. \textbf{\underline{RL+Hybrid2Seq}} \cite{DBLP:conf/kbse/WanZYXY0Y18} incorporates ASTs and code sequences into a deep reinforcement learning framework, while \textbf{\underline{DeepCom}} \cite{DBLP:conf/iwpc/HuLXLJ18} encodes the node sequences traversed from ASTs to capture the structural information. \textbf{\underline{API+Code}} \cite{DBLP:conf/ijcai/HuLXLLJ18} involves API knowledge in the code summarization procedure. \textbf{\underline{Dual model}} \cite{DBLP:conf/kbse/Wei19} adopts a dual learning framework to exploit the duality of code summarization and code generation tasks. One of the most recent approaches, denoted as \textbf{\underline{NeuralCodeSum}}~\cite{DBLP:conf/acl/AhmadCRC20}, which integrates the \textbf{\underline{vanilla Transformer}} \cite{DBLP:conf/nips/VaswaniSPUJGKP17} with relative position encoding (RPE) and copy attention. Another recent approach \textbf{\underline{Transformer+GNN}} \cite{DBLP:conf/acl/ChoiBNL21} applies graph convolution to obtain structurally-encoded node representations and passes sequences of the graph-convolutioned AST nodes into Transformer.

We also compare our approach with relational Transformers~\cite{DBLP:conf/iclr/HellendoornSSMB20,DBLP:conf/iclr/ZugnerKCLG21} which involve structural information for code representation learning. \textbf{\underline{GREAT}~\cite{DBLP:conf/iclr/HellendoornSSMB20}} biases vanilla Transformers with relational information from graph edge types. \textbf{\underline{CodeTransformer}~\cite{DBLP:conf/iclr/ZugnerKCLG21}} focuses on multilingual code summarization and proposes to build upon language-agnostic features such as source code and AST-based features.

During implementation, we either directly copy the results claimed in the original papers or reproduce the results strictly following the released repositories for most baselines except for GREAT and CodeTransformer. 
For GREAT~\cite{DBLP:conf/iclr/HellendoornSSMB20}, 12 types of information including control flow graph and syntactic features are adopted for model training, as 
no replication package is available. Due to the difficulty of complete replication, we follow the strategy in Z{\"{u}}gner et al.'s work \cite{DBLP:conf/iclr/ZugnerKCLG21} by employing 
the same structural information as \tool during replication. For CodeTransformer, although a replication package is provided by the authors, not all the benchmark data can be successfully preprocessed. For Java, only 61,250 of 69,708 code snippets in the training set, 7,621 of 8,714 in the validation set, and 7,643 of 8,714 in the test set pass the preprocessing step; while for Python, all the code snippets can be well preprocessed. To ensure 
the consistency of evaluation data, we compare \tool with CodeTransformer on the Java dataset separately. We use the same model settings for implementing CodeTransformer, including the layer number, head number, etc.

\subsection{Parameter Settings}
\tool is composed of $8$ layers and $8$ heads in its Transformer architecture and the hidden size of the model is 512. We use Adam optimizer with the initial learning rate set to $10^{-4}$, batch size set to $32$, and dropout rate set to $0.2$ during the training. We train our model for at most $200$ epochs and select the checkpoint with the best performance on the validation set for further evaluation on the test set. We report the performance of \tool and each ablation experiment by running three times and taking the average. To avoid over-fitting, we early stop the training if the performance on the validations set does not increase for $20$ epochs. For the control factors of heads distribution and data flow, we set them to $1$ and $5$, respectively. We will discuss 
{optimal} parameters selection in Section~\ref{subsec:param}. Our experiments are conducted on a single Tesla V100 GPU for about 30 hours, and we train our model from scratch.
\section{Experimental Results}\label{sec:results}

\begin{table}[t]
\centering
\caption{Comparison results with baseline models. The \textbf{bold} figures indicate the best results. ``*'' denotes statistical significance in comparison to the baseline models we reproduced (i.e., two-sided $t$-test with $p$-value$<0.01$).}\label{tab:results}
\aboverulesep=0ex
\belowrulesep=0ex
\scalebox{1.0}{
\begin{tabular}{l|ccc|ccc}
\toprule
\multirow{2}{*}{\textbf{Approach}} & \multicolumn{3}{c|}{\textbf{Java}} & \multicolumn{3}{c}{\textbf{Python}} \\
\cmidrule{2-7} 
& BLEU-4 & METEOR & ROUGE-L &  BLEU-4 & METEOR & ROUGE-L\\
\midrule
{ CODE-NN \citep{DBLP:conf/acl/IyerKCZ16} } & 27.60 & 12.61 & 41.10 & 17.36 & 09.29 & 37.81 \\
{ Tree2Seq \citep{DBLP:conf/acl/EriguchiHT16} } & 37.88 & 22.55 & 51.50 & 20.07 & 08.96 & 35.64 \\
 { RL+Hybrid2Seq \citep{DBLP:conf/kbse/WanZYXY0Y18}} & 38.22 & 22.75 & 51.91 & 19.28 & 09.75 & 39.34 \\
 { DeepCom \citep{DBLP:conf/iwpc/HuLXLJ18}} & 39.75 & 23.06 & 52.67 & 20.78 & 09.98 & 37.35 \\
{ API+Code \citep{DBLP:conf/ijcai/HuLXLLJ18} } & 41.31 & 23.73 & 52.25 & 15.36 & 08.57 & 33.65 \\
{ Dual Model \citep{DBLP:conf/nips/WeiL0FJ19} } & 42.39 & 25.77 & 53.61 & 21.80 & 11.14 & 39.45 \\
{ Code2seq\citep{DBLP:conf/iclr/AlonBLY19}  } &  12.19&  08.83&  25.61&  18.69&  13.81&  34.51\\
\midrule
{ Vanilla Transformer \citep{DBLP:conf/nips/VaswaniSPUJGKP17} } & 44.20 & 26.83 & 53.45 & 31.34 & 18.92 & 44.39 \\
{ NeuralCodeSum \citep{DBLP:conf/acl/AhmadCRC20} } & 45.15 & 27.46 & 54.84 & 32.19 & 19.96 & 46.32 \\
{ GREAT\citep{DBLP:conf/iclr/HellendoornSSMB20}  } & 44.97 & 27.15 & 54.42 & 32.11 & 19.75 & 46.01 \\
{ CodeTransformer \citep{DBLP:conf/iclr/ZugnerKCLG21}  }  & -- & -- & -- & 27.63 & 14.29 & 39.27 \\
{ Transformer+GNN\citep{DBLP:conf/acl/ChoiBNL21} } & 45.49  & 27.17 &  54.82 & 32.82  &  20.12 & 46.81\\
\midrule
{ \textbf{\tool} } &\textbf{45.89}* & \textbf{27.85}* & \textbf{55.79}* & \textbf{33.04}* & \textbf{20.52}* & \textbf{47.01}*  \\
\bottomrule
\end{tabular}
}
\end{table}

\begin{table}[t]
\centering
\caption{Comparison results with CodeTransformer on the dataset preprocessed by CodeTransformer. The \textbf{bold} figures indicate the best results. ``*'' denotes statistical significance in comparison to the baseline models (i.e., two-sided $t$-test with $p$-value$<0.01$).}\label{tab:results_codetrans_1}
\aboverulesep=0ex
\belowrulesep=0ex
\scalebox{1.0}{
\begin{tabular}{l|ccc}
\toprule
\multirow{2}{*}{\textbf{Approach}} & \multicolumn{3}{c}{\textbf{Java}} \\
\cmidrule{2-4} 
& BLEU-4 & METEOR & ROUGE-L\\
\midrule
{ CodeTransformer \citep{DBLP:conf/iclr/ZugnerKCLG21}  }  & 39.81 & 24.22 & 51.96 \\
{ \textbf{\tool} } &\textbf{44.59}* & \textbf{27.32}* & \textbf{54.41}*  \\
\bottomrule
\end{tabular}
}
\end{table}

In this section, we elaborate on the comparison results with the baselines to evaluate \tool's capability in accurately generating code summaries. Our experiments are aimed at answering the following research questions:

\begin{enumerate}[label=\bfseries RQ\arabic*:,leftmargin=.5in]
    \item What is the performance of \tool in code summary generation?
    
    \item What is the impact of the involved code structural properties and the design of hierarchical attentions on the model performance? 

    
    \item How accurate is \tool under different parameter settings?

\end{enumerate}

\subsection{Answer to RQ1: Comparison with the Baselines}

The experimental results on the benchmark datasets are shown in Table~\ref{tab:results}.
For the vanilla Transformer and the NeuralCodeSum \cite{DBLP:conf/acl/AhmadCRC20}, we reproduce their experiments under the same hyper-parameter settings as the Transformer in \tool to ensure fair comparison. We compare \tool with CodeTransformer~\cite{DBLP:conf/iclr/ZugnerKCLG21} on the Java dataset separately, in which both approaches are trained and evaluated on the same dataset, with results shown in Table~\ref{tab:results_codetrans_1}. Based on Table~\ref{tab:results} and Table~\ref{tab:results_codetrans_1}, we {summarize} the following findings:

\textbf{Code structural properties are beneficial for source code summarization.} Comparing Tree2Seq/DeepCom with CODE-NN, we can find that the structure information brings a great improvement in the performance. For example, both Tree2Seq and DeepCom outperform CODE-NN by at least 37.2\%, 78.8\%, and 25.3\% respectively regarding the three metrics on the Java dataset. Although {no consistent improvement across all metrics is observed} on the Python dataset, Tree2Seq/DeepCom {still} shows an obvious increase on the BLEU-4 metric.

\textbf{Transformer-based approaches perform better than RNN-based approaches.} The four Transformer-based approaches~\cite{DBLP:conf/iclr/HellendoornSSMB20,DBLP:conf/iclr/AlonBLY19,DBLP:conf/acl/AhmadCRC20,DBLP:conf/nips/VaswaniSPUJGKP17,DBLP:conf/acl/ChoiBNL21} outperform all the other baselines, 
{with} NeuralCodeSum~\cite{DBLP:conf/acl/AhmadCRC20} {giving} 
better performance {compared to the vanilla Transformer}. The vanilla Transformer already achieves better performance than the top seven RNN-based approaches {with various types of structural information being incorporated, showing}
the efficacy of Transformer for the task.  
On the Python dataset, NeuralCodeSum {outperforms} 
the best RNN-based baseline, Dual Model~\cite{DBLP:conf/kbse/Wei19}, by 47.7\% and 17.4\% in terms of the BLEU-4 and ROUGE-L metrics, respectively.

\textbf{The proposed \tool is effective in code summarization.} Comparing \tool with the NeuralCodeSum and Transformer+GNN, \tool achieves 
the best results on both benchmark datasets, yet without introducing any extra model parameters. Specifically, \tool improves the best baseline by 1.7\% and 0.4\% in terms of ROUGE-L score on the Java and Python dataset, respectively.

\textbf{The combination of the structural information in \tool is effective.} By comparing \tool with other Transformer models with structural information involved such as GREAT~\cite{DBLP:conf/iclr/HellendoornSSMB20}, CodeTransformer~\cite{DBLP:conf/iclr/ZugnerKCLG21} and Transformer+GNN~\cite{DBLP:conf/acl/ChoiBNL21},
\tool achieves the best results on both benchmark datasets.
Specifically, \tool improves the best baseline by 2.5\% and 2.0\% in terms of METEOR score on the Java and Python dataset, respectively.

\begin{table*}[t]
\centering
\caption{Ablation study on different part of our model. The \textbf{bold} figures indicate the best results. ``*'' denotes statistical significance in comparison to the baseline models we reproduced (i.e., two-sided $t$-test with $p$-value$<0.01$).}\label{tab:ablation}
\aboverulesep=0ex
\belowrulesep=0ex
\scalebox{1.0}{
\begin{tabular}{l|ccc|ccc}
\toprule
\multirow{2}{*}{\textbf{Approach}} & \multicolumn{3}{c|}{\textbf{Java}} & \multicolumn{3}{c}{\textbf{Python}} \\
\cmidrule{2-7} 
& BLEU-4 & METEOR & ROUGE-L &  BLEU-4 & METEOR & ROUGE-L\\
\midrule
{ \tool w/o token info.} & 44.92 & 27.35 & 54.69 & 32.18 & 19.87 & 46.14 \\
{ \tool w/o statement info.} & 44.61 & 27.08 & 54.04 & 32.26 & 19.66 & 46.08 \\
{ \tool w/o data flow info.} & 45.52 & 27.65 & 55.40 & 32.58 & 20.16 & 46.57 \\
{ \tool w/o hierarchical attention} & 45.53 & 27.72 & 55.48 & 32.93 & 20.38 & 46.73 \\
{ \tool w/o copy attention} & 45.24 & 27.49 & 55.01 & 31.89 & 19.26 & 45.31 \\
{ SG-Trans\textsubscript{soft} } &45.37 & 27.65 & 55.09 &  32.77&  19.96& 46.74\\
{ \textbf{\tool} } &\textbf{45.89}* & \textbf{27.85}* & \textbf{55.79}* & \textbf{33.04}* & \textbf{20.52}* & \textbf{47.01}*  \\
\bottomrule
\end{tabular}
}
\end{table*}


\subsection{Answer to RQ2: Ablation Study}\label{subsec:ablation}
We further perform ablation studies to validate the impact of the involved code structural properties and the hierarchical structure-variant attention approach. Besides, to evaluate the efficacy of the hard mask attention mechanism for combining token-level and statement-level information in \tool, we create a comparative approach, named as SG-Trans\textsubscript{soft}, by changing the hard mask into soft mask. Specifically, for SG-Trans\textsubscript{soft}, we follow NeuralCodeSum~\cite{DBLP:conf/acl/AhmadCRC20}, and only add the relative position embedding for subtoken pairs $x_i$ and $x_j$ if they are in the same token or statement. The results are shown in Table~\ref{tab:ablation}.

\textbf{Analysis of the involved code structure.} We find that all the three structure types, including code token, statement and data flow, contribute to the model performance {improvement} but with varied degrees. Specifically, local syntactic structures play a more important role
than the global data flow structure. For example, removing the statement information leads to a significant performance drop at around 2.8\% and 2.4\% regarding the BLEU-4 score. This suggests the importance of modeling the semantic relations among tokens of the same statement for code summarization. With the data flow information eliminated, \tool also suffers from a performance drop, which may indicate that the data dependency relations are hard to be learnt by Transformer implicitly.

\textbf{Analysis of the hierarchical structure-variant attention mechanism.} We replace the hierarchical structure-variant attention with uniformly-distributed attention, i.e., $\Omega^l=[\omega_t^l, \omega_s^l, \omega_f^l, \omega_o^l]=[2, 2, 2, 2]$, for the ablation analysis. As can be found in Table~\ref{tab:results}, without the hierarchical structure design, the model's performance decreases 
on {all metrics for both datasets}. The results demonstrate the positive impact of the hierarchical structure-variant attention mechanism.

\textbf{Analysis of the copy attention.} As shown in Table~\ref{tab:results}, excluding the copy attention results in a significant drop to \tool's performance, similar to what has been observed in Ahmad et al.’s work~\citep{DBLP:conf/acl/AhmadCRC20}. This shows that the copy attention is useful for alleviating the OOV issue and facilitating {better} code summarization.

\textbf{Analysis of the hard mask attention mechanism.} 
As shown in Table~\ref{tab:ablation}, we can find that \tool performs constantly better than SG-Trans\textsubscript{soft} on both Java and Python datasets with respect to all the metrics. For example, on Java dataset replacing hard-mask with soft mask leads to a performance drop at 1.1\% and 1.3\% in terms of the BLEU-4 and ROUGE-L metrics, respectively, which indicates that the hard mask attention is effective at capturing the local information.

\begin{figure*}[ht]
    \centering
    \begin{subfigure}[b]{0.48\textwidth}
        \includegraphics[width=\textwidth]{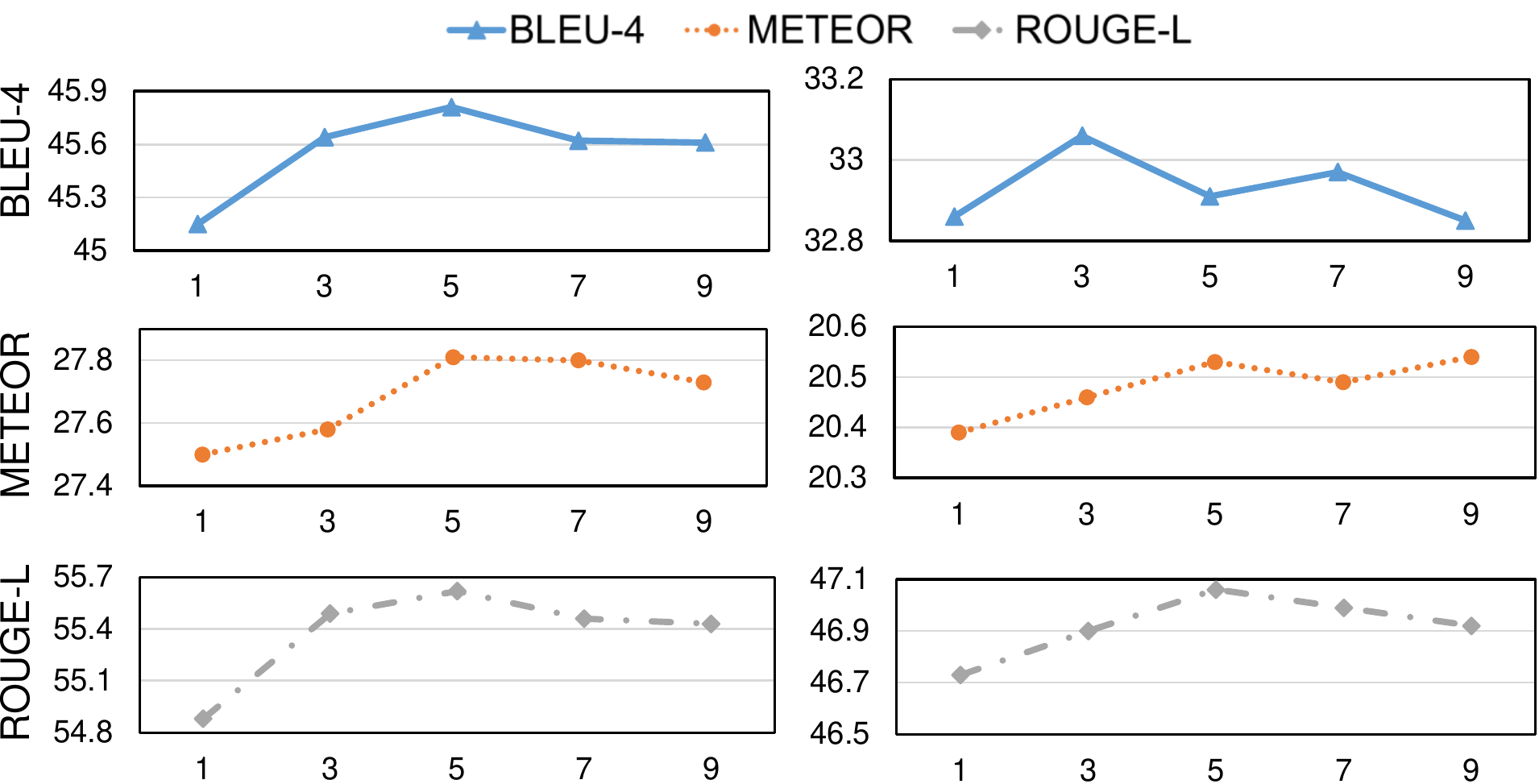}
        \caption{Analysis of the parameter $\mu$.}
        \label{tab:alpha}
      \end{subfigure}
      \hfill
      \begin{subfigure}[b]{0.48\textwidth}
        \includegraphics[width=\textwidth]{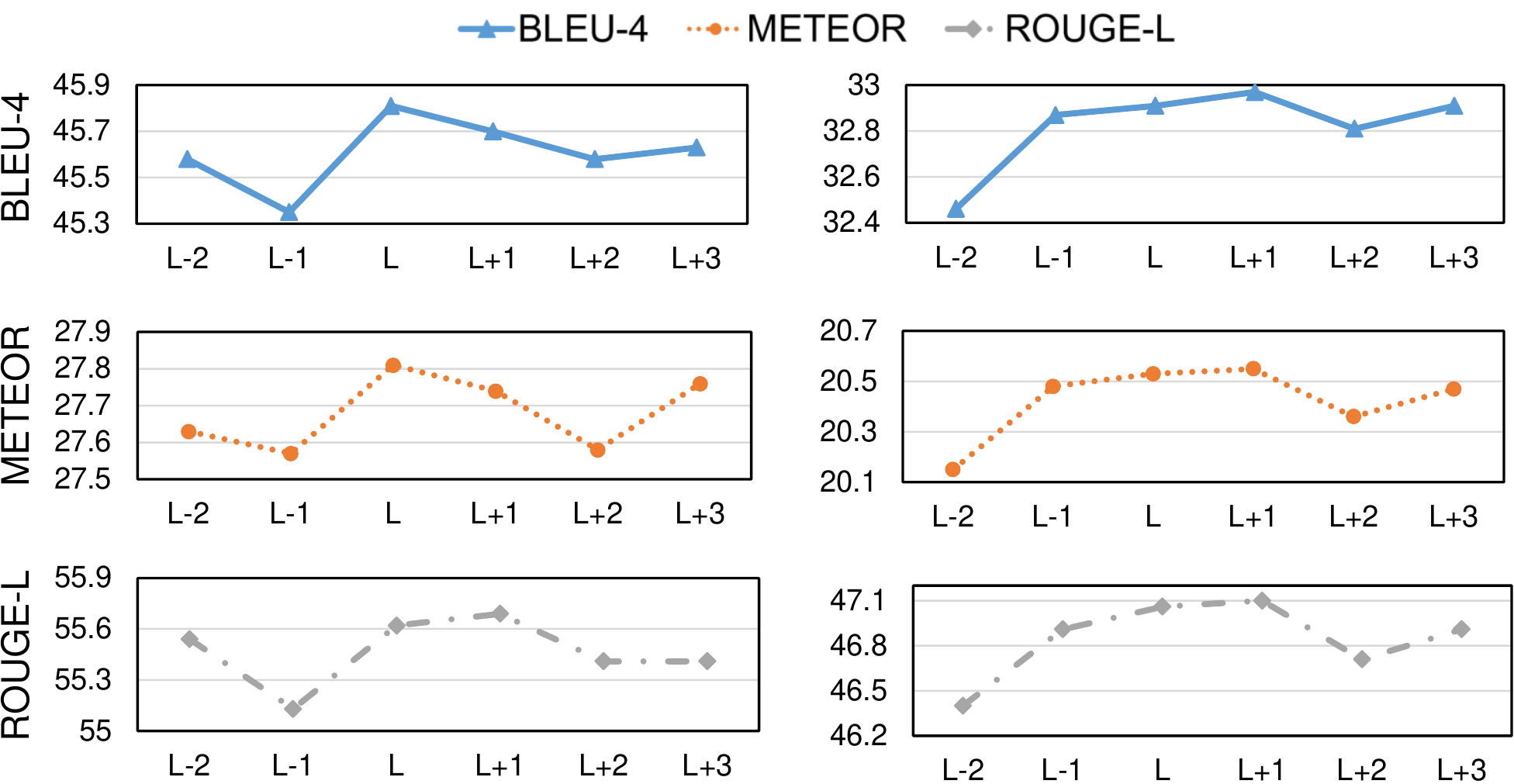}
        \caption{Analysis of the parameter $k$. $L=8$.}
        \label{tab:k}
      \end{subfigure}
    \caption{Influence of the hyper-parameters $\mu$ and $k$ on the model performance.}
	\label{fig:params}
\end{figure*}

\subsection{Answer to RQ3: Parameter Sensitivity Analysis}\label{subsec:param}
In this section we analyze the impact of two key hyper-parameters on the model performance, i.e., the control factor $\mu$ for adjusting the integration degree of the data flow structure and the parameter $k$ to control the head distribution.

\textbf{The parameter $\mu$.} Figure~\ref{fig:params} (a) shows the performance variation with the changes of $\mu$ while keeping other hyper-parameters fixed. For the Java dataset, the model achieves the best scores when $\mu=5$. Lower or higher parameter values do not give better results. While for the Python dataset, a similar trend is observed for the BLEU-4 and ROUGE-L metrics where the model performs the best when $\mu$ equals to 3 and 5, respectively. In this work, we set $\mu$ to 5 since the model can produce relatively better results on both datasets.

\textbf{The parameter $k$.} We observe the performance changes when the control factor $k$ of the head distribution takes values centered on layers of \tool $L$. Figure~\ref{fig:params} (b) illustrates the results. We can find that \tool can well balance the distribution of local and global structure-guided head attention when $k=L$ or $k=L+1$. As $k$ gets larger, \tool would be more biased by the local structure and tend to generate inaccurate code summary. In our work here, we set $k=L$. 

\subsection{Human Evaluation}

In this 
section, we perform human evaluation to qualitatively evaluate the summaries generated by four Transformer-based baselines, including vanilla Transformer, NeuralCodeSum, GREAT and CodeTransformer, and also our model \tool. We do not involve the baseline Transformer+GNN, due to the lack of replication package.
The human evaluation is conducted through online questionnaire. In total, 10 software developers are invited for evaluation. All participants
have programming experience in software development for at least four years, and none of them is a co-author of the paper.
Each participant is invited to read 60 code snippets and judge the quality of summaries generated by vanilla Transformer, NeuralCodeSum, CodeTransformer, GREAT and \tool. Each of them will be paid 30 USD upon completing the questionnaire.

\begin{figure}[t]
\centering
\includegraphics[width=0.7\textwidth]{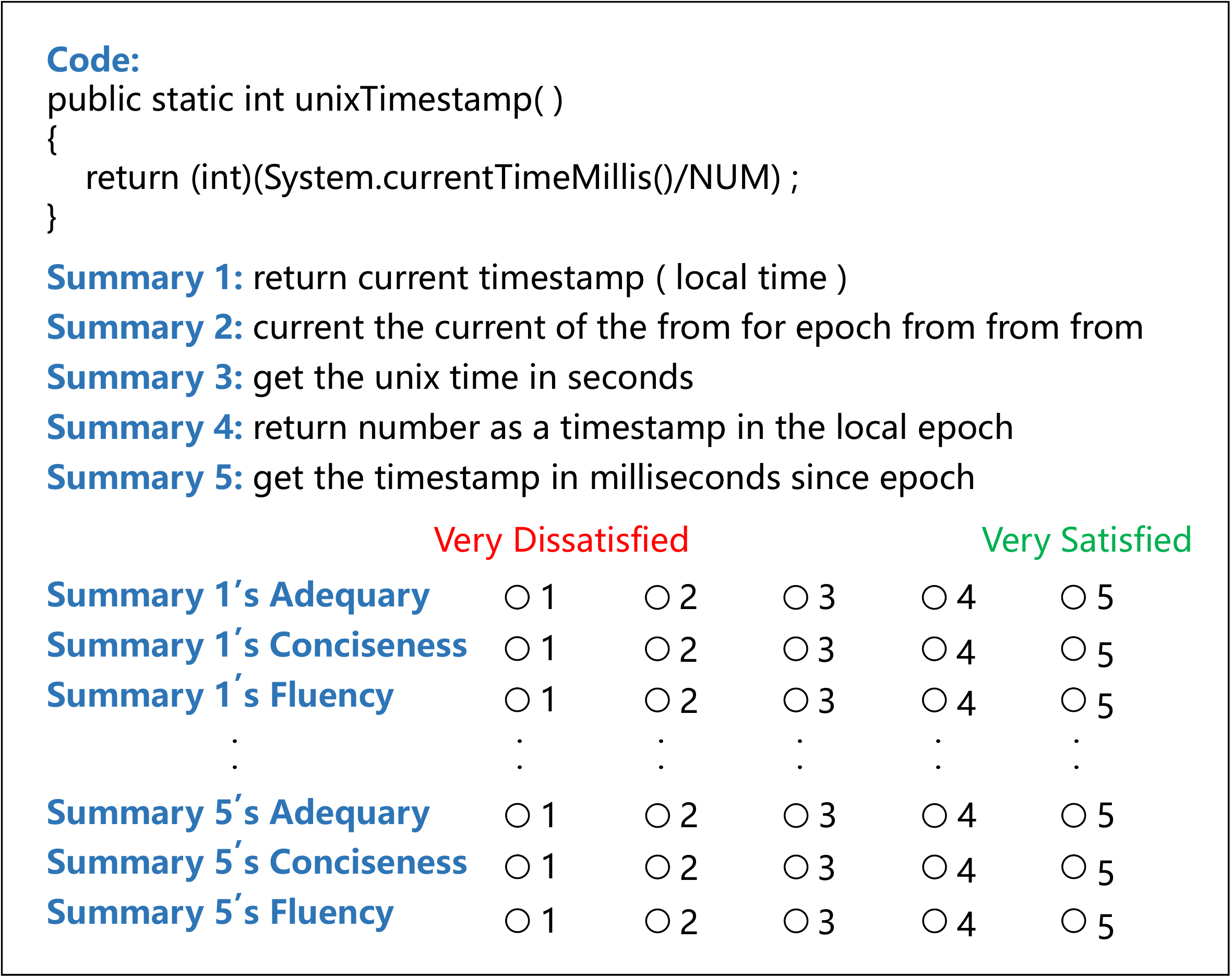}
\caption{An example of questions in our questionnaire. The two-dot symbols indicate the simplified rating schemes for Summary 2,3,4.}
\label{fig:survey_example}
\end{figure}

\begin{table}[t]
\centering
\aboverulesep=0ex
\belowrulesep=0ex
\caption{Human evaluation results. The \textbf{bold} figures indicate the best results.}\label{tab:human_study}
\scalebox{0.93}{
\begin{tabular}{l|c|ccccc}
\toprule
Dataset & Metrics & Transformer & CodeTransformer & NeuralCodeSum & GREAT & \tool \\
\midrule
\multirow{3}{*}{\textbf{\ Java}} & Adequacy & 3.35 &2.67 & 3.28 & 3.44 & \textbf{3.65}\\
& Conciseness & 4.20 & 3.49 & 4.32 & 4.36 & \textbf{4.50}\\
& Fluency & 4.32 & 3.25 & 4.36 & 4.50 & \textbf{4.59}\\
\midrule
\multirow{3}{*}{\textbf{\ Python}} & Adequacy & 2.61 & 1.92 & 3.04 & 2.83 & \textbf{3.21}\\
& Conciseness & 3.84 & 2.62 & 4.01 & 4.05 & \textbf{4.21}\\
& Fluency & 4.06 & 2.39 & 4.21 & 4.26 & \textbf{4.33}\\
\bottomrule
\end{tabular}
}
\end{table}

\subsubsection{Survey Design}

We randomly selected 200 code snippets, with 100 in Java and 100 in Python, for evaluation. As shown in Figure~\ref{fig:survey_example}, in the questionnaire, each question comprises a code snippet and summaries generated by the five models. Each participant will be given 60 questions and each question will be evaluated by three different participants. For each question, the summaries generated by the models are randomly shuffled to eliminate the order bias.

The quality of the generated summaries is evaluated in three aspects, including \emph{Adequacy}, \emph{Conciseness}, and \emph{Fluency}, with the 1-5 Likert scale (5 for excellent, 4 for good, 3 for acceptable, 2 for marginal, and 1 for poor). We explained the meaning of the three evaluation metrics at the beginning of the questionnaire: The metric ``adequacy” measures how much the functional meaning of the code is preserved after summarization; the metric ``conciseness” measures the ability to express the function of code snippet without unnecessary words; while the metric ``fluency” measures the quality of the generated language such as the correctness of grammar. 

\begin{figure}[t]
     \centering
     \begin{subfigure}[h]{0.45\textwidth}
        \centering
    	\includegraphics[width=0.8 \textwidth]{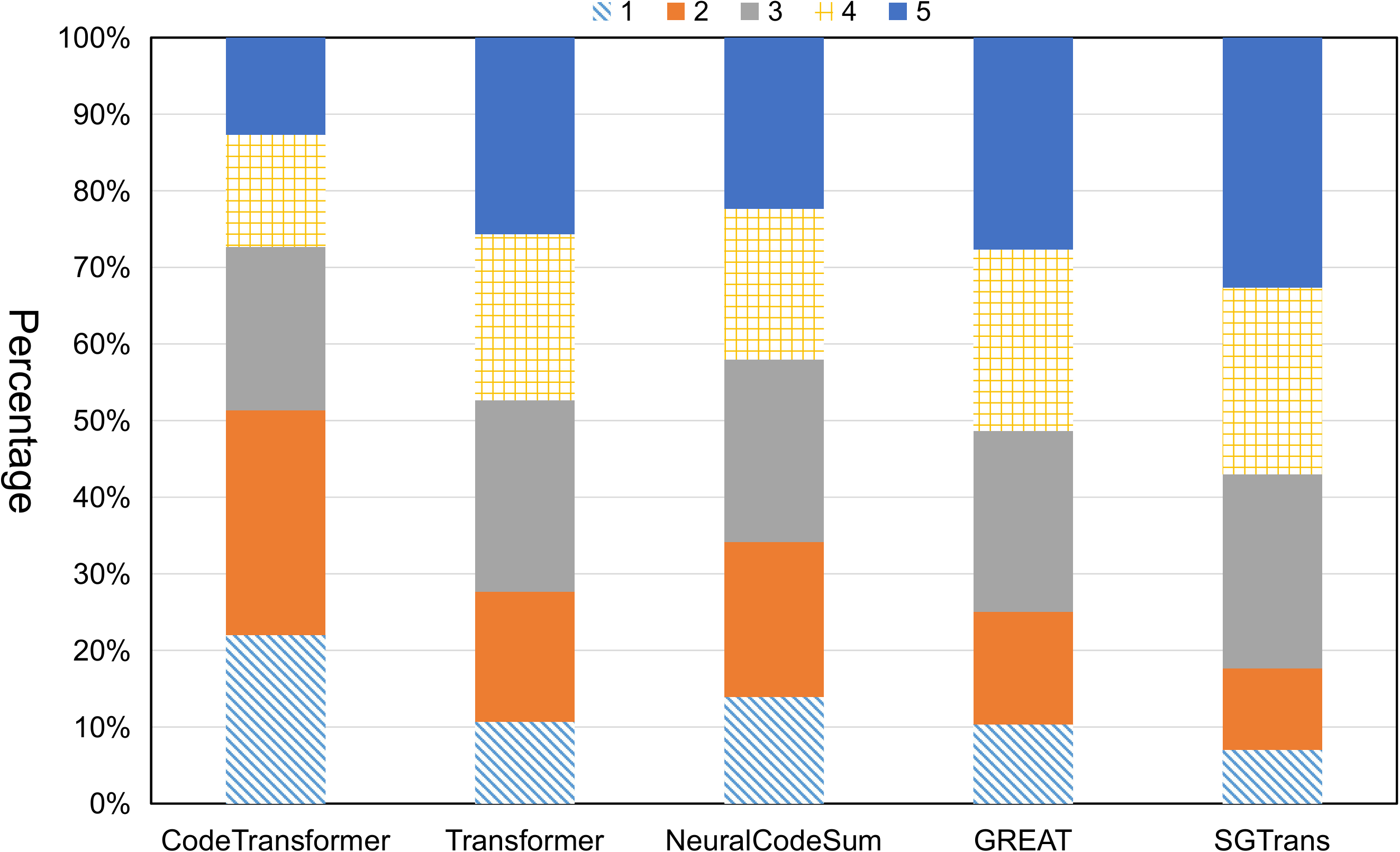}
    	\caption{\small Adequacy metric for the Java dataset.}
        \end{subfigure}
        \hfill
        \begin{subfigure}[h]{0.45\textwidth}
        \centering
        \includegraphics[width=0.8 \textwidth]{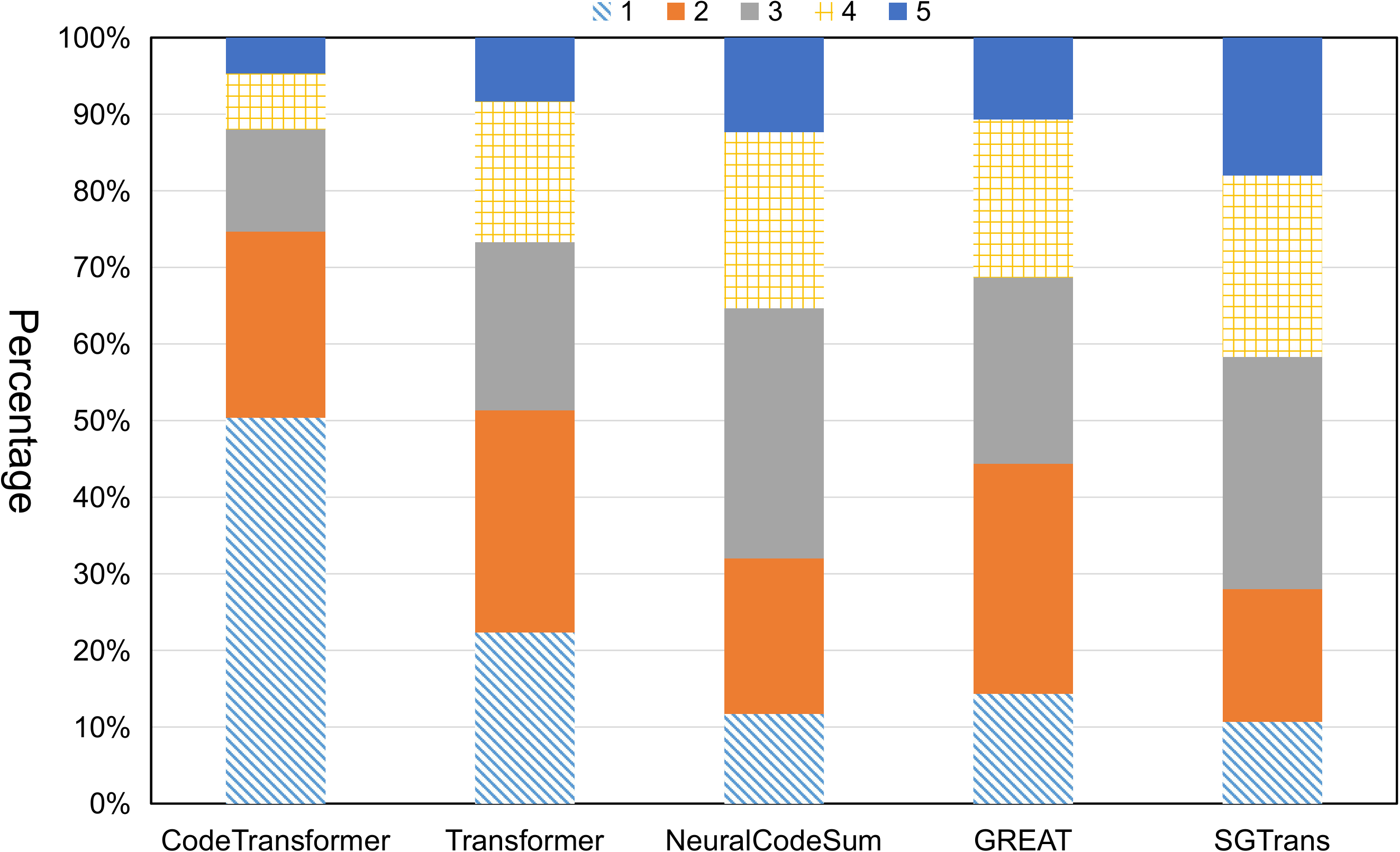}
        \caption{\small Adequacy metric for the Python dataset.}
        \end{subfigure}
        \begin{subfigure}[h]{0.45\textwidth}
        \vspace*{0.15cm}
        \centering
        \includegraphics[width=0.8 \textwidth]{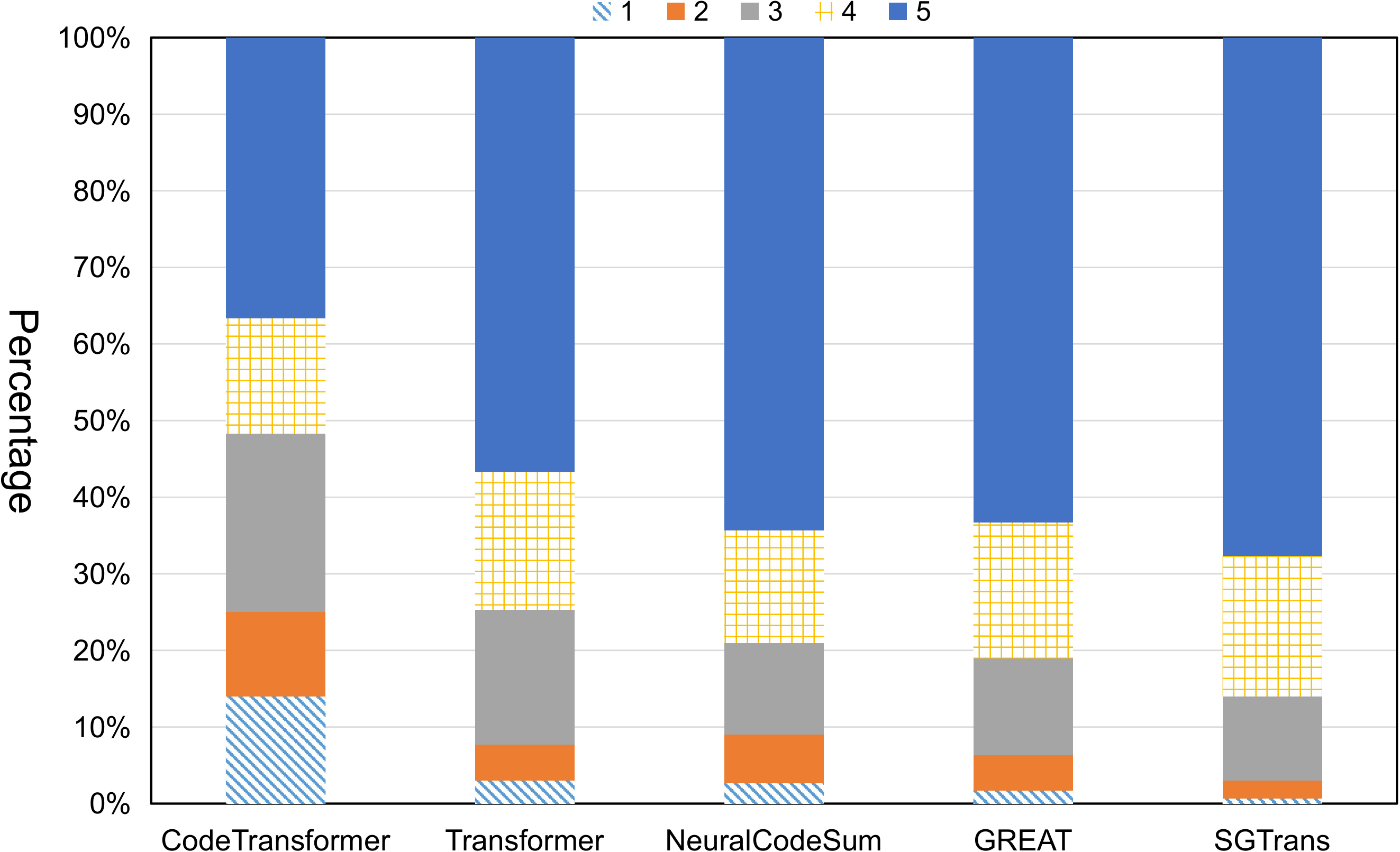}
        \caption{\small Conciseness metric for the Java dataset.}
        \end{subfigure}
        \hfill
        \begin{subfigure}[h]{0.45\textwidth}
        \vspace*{0.15cm}
        \centering
        \includegraphics[width=0.8 \textwidth]{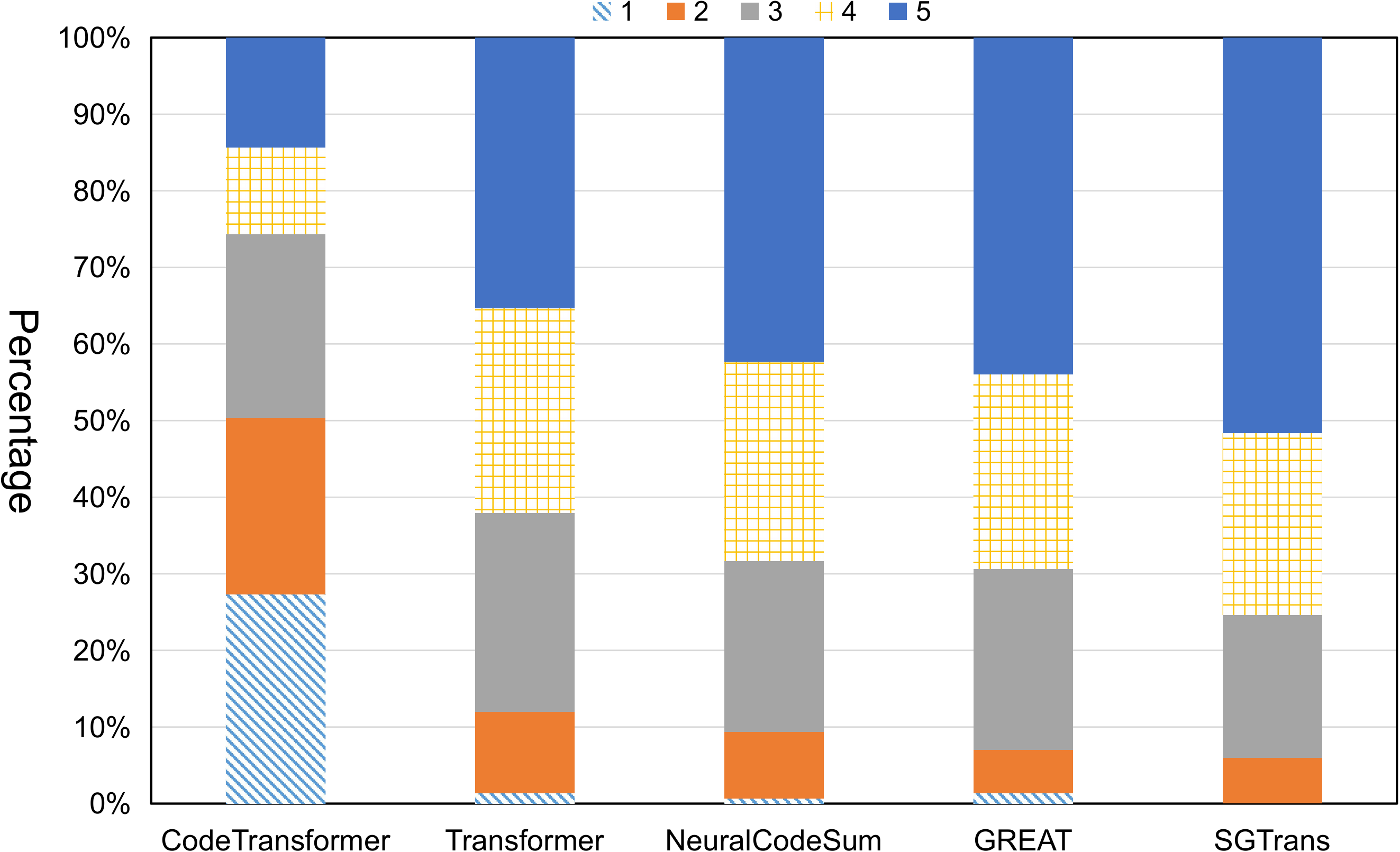}
        \caption{\small Conciseness metric for the Python dataset.}
        \end{subfigure}
        \begin{subfigure}[h]{0.45\textwidth}
        \vspace*{0.15cm}
        \centering
        \includegraphics[width=0.8 \textwidth]{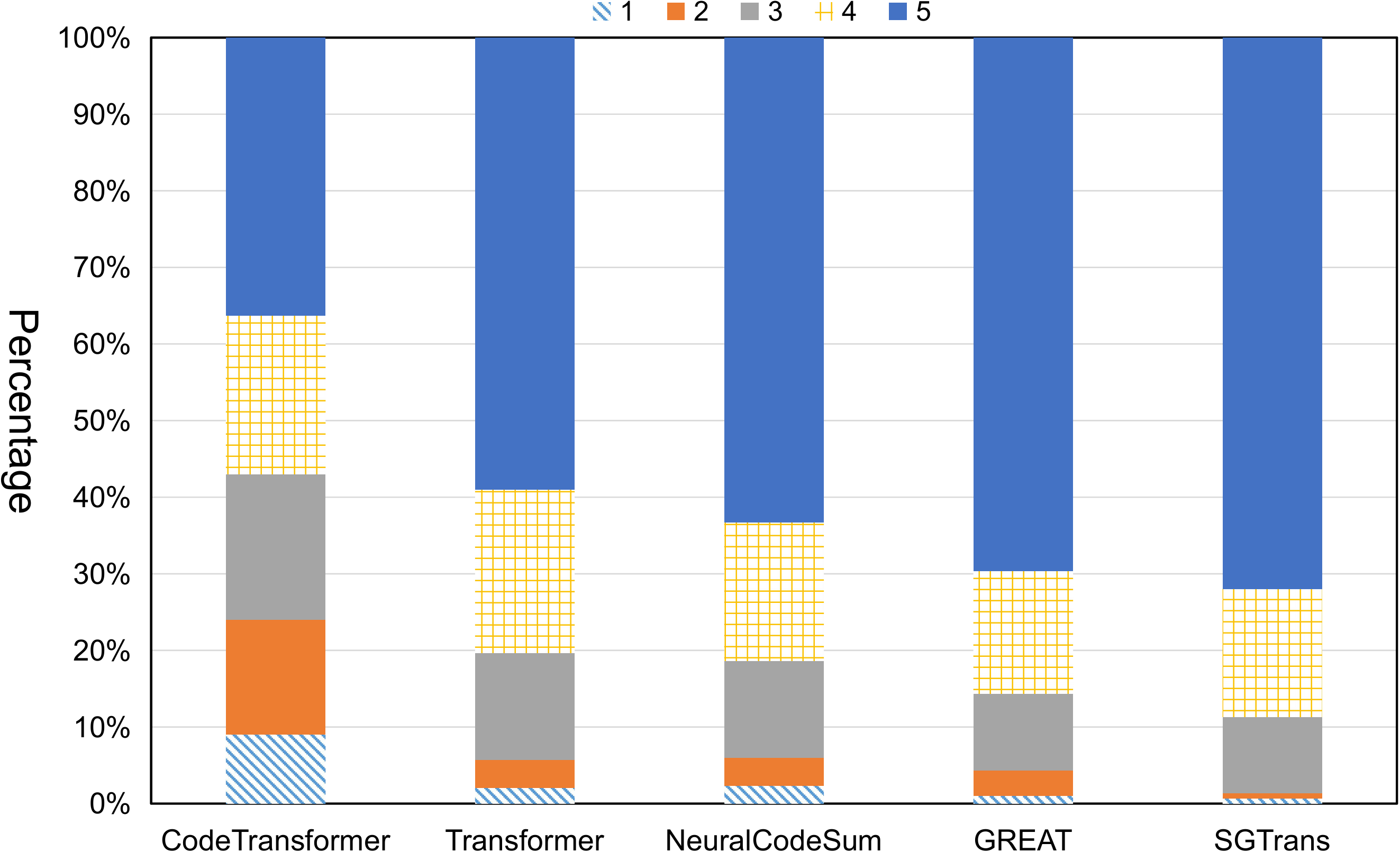}
        \caption{\small Fluency metric for the Java dataset.}
        \end{subfigure}
        \hfill
        \begin{subfigure}[h]{0.45\textwidth}
        \vspace*{0.15cm}
        \centering
        \includegraphics[width=0.8 \textwidth]{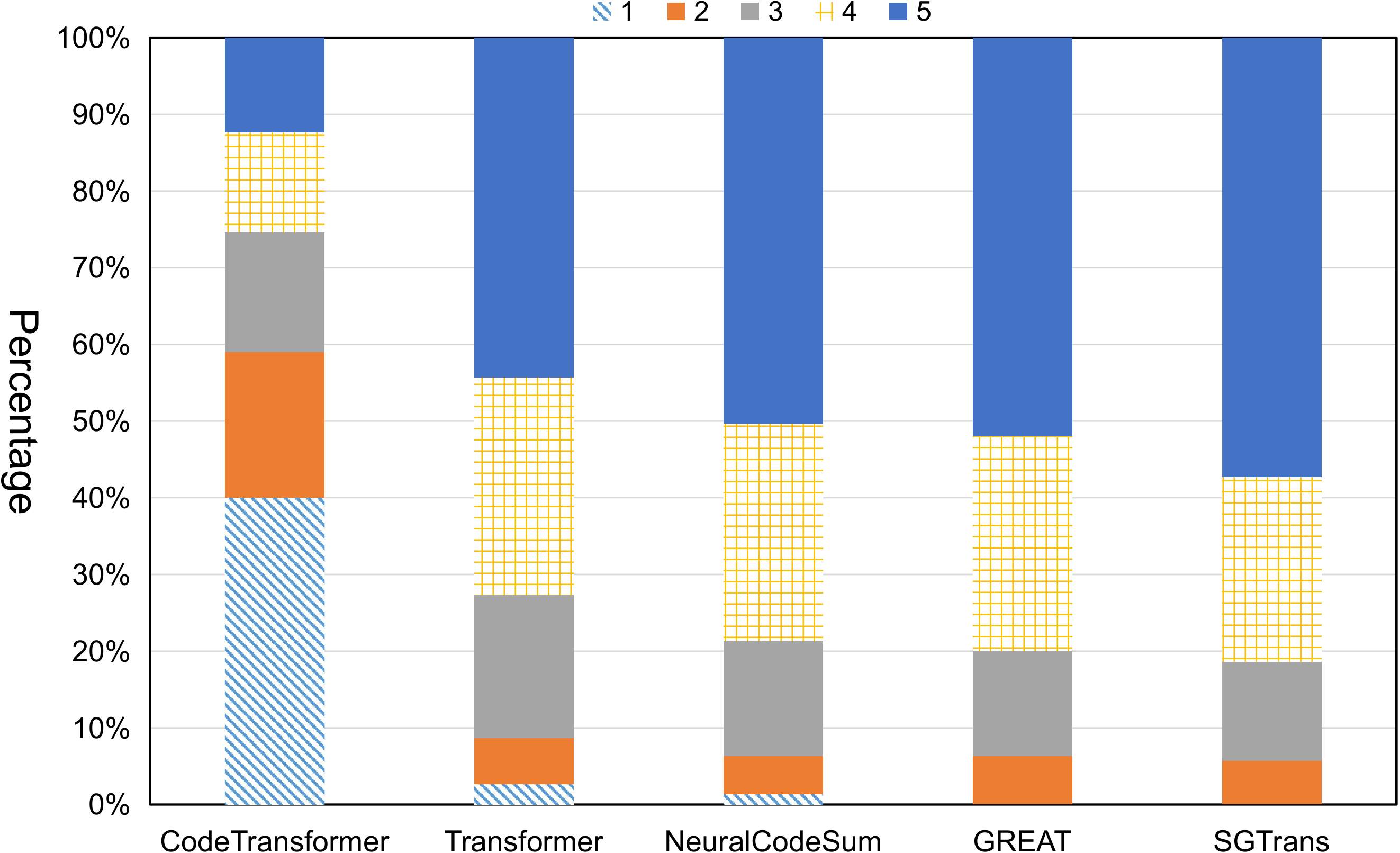}
        \caption{\small Fluency metric for the Python dataset.}
        \end{subfigure}
        
         \caption{Distribution of the rating scores in human evaluation on the two datsets. The ``Transformer'' on the horizontal axis denotes the ``vanilla Transformer'' approach.}
     \label{fig:human}
\end{figure}

\subsubsection{Survey Design}

We finally received 600 sets of scores with 3 sets of scores for each code-summary pair from the human evaluation. On average, the participants spent 2 hours on completing the questionnaire, with the median completion time at 1.67 hours. The inter-annotator agreement of the two sets is evaluated with the widely-used metric Cohen's kappa. The average Cohen's kappa scores for the Java and Python datasets are 0.66 and 0.58
, respectively, indicating that the participants achieved at least moderate agreement on both datasets.

The evaluation results are illustrated in Table~\ref{tab:human_study} and Figure~\ref{fig:human}. We find that the summaries generated by \tool receive the highest scores on both datasets and with respect to all the metrics. For the Java dataset, as shown in Table~\ref{tab:human_study}, \tool improves the baseline models by at least 6.1\%, 3.2\% and 2.0\% with respect to the adequacy, conciseness, and fluency metrics, respectively. As can be observed from Figure~\ref{fig:human} (a), (c) and (e), summaries generated by \tool receive the most 5-star ratings and fewest 1/2-star ratings from the participants, comparing with the summaries produced by other models for each 
metric. Specifically, regarding the fluency metric, only 1.3\% of the participants gave 1/2-star ratings to the summaries generated by \tool, while other approaches receive at least 6.0\% 1/2-star ratings and CodeTransformer receives even 24.0\%. The score distributions indicate that \tool better captures the functionality of given code snippets and generates nature language comments with higher quality.

For the Python dataset, as shown in Table~\ref{tab:human_study}, NeuralCodeSum and GREAT significantly outperform the vanilla Transformer and CodeTransformer; while \tool is more powerful, further boosting the best baseline approach by 5.6\%, 4.0\% and 1.6\% in terms of the adequacy, conciseness, and fluency, respectively. As can be observed from Figure~\ref{fig:human} (b), (d) and (f), summaries generated by \tool receive the most 5-star ratings and fewest 1/2-star ratings from the annotators. Specifically, regarding the adequacy metric, 18.0\% of the participants gave 5-star ratings to the summaries generated by \tool, with only 4.6\% for the CodeTransformer approach, 12.3\% for the strongest baseline NeuralCodeSum. For the conciseness metric, 51.7\% of the participants gave 5-star ratings to the summaries generated by \tool and the two best baseline approaches NeuralCodeSum and GREAT only receive 42.3\% and 44.0\% 5-star ratings, respectively. The score distributions indicate that the summaries generated by \tool can better describe the function of code snippets and have a more concise and accurate expression.

\subsection{Further Evaluation on Generated Summaries}

\begin{table}[t]
\centering
\caption{The quality of summaries generated by \tool and humans. The term ``auto-generated'' indicates the summaries output by \tool.}\label{tab:results_rating_write}
\aboverulesep=0ex
\belowrulesep=0ex
\scalebox{0.95}{
\begin{tabular}{l|l|ccc|ccc}
\toprule
\multirow{2}{*}{\textbf{Predicted Summary}} & \multirow{2}{*}{\textbf{Test}} & \multicolumn{3}{c|}{\textbf{Java}} & \multicolumn{3}{c}{\textbf{Python}} \\
\cmidrule{3-8} 
& & BLEU-4 & METEOR & ROUGE-L &  BLEU-4 & METEOR & ROUGE-L\\
\midrule
{ Auto-generated} & {Reference } &77.06 & 51.67 & 89.95 &  75.09&  49.04& 87.56\\
{ Human-generated } & {Reference } &19.04 & 13.33 & 32.83 &  23.63&  19.82& 38.95\\
\bottomrule
\end{tabular}
}
\end{table}

\begin{table*}[t]
\newcommand{\tabincell}[2]{\begin{tabular}{@{}#1@{}}#2\end{tabular}}
\centering
\caption{The code snippets that cannot be understood by the annotators.}\label{tab:human_case_study_fail}
\scalebox{0.9}{
\begin{tabular}{l}
\toprule
 \textbf{Example (1) in Python:} \\
 \tabincell{l}{
 \texttt{\textcolor[RGB]{59,162,59}{\textbf{def}} \textcolor{blue}{poly\_TC}(f, K): }\\ \hspace{2em} \texttt{if (not f):} \\ \hspace{4em} \texttt{return K.zero} \\ \hspace{2em} \texttt{else:} \\ \hspace{4em} \texttt{return f[(-1)]} \\
 \textbf{Human-generated}: \textcolor[RGB]{255,0,0}{\textbf{cannot understand}}  \\  \textbf{\tool}: return trailing coefficient of f \\ \textbf{Ground truth}: return trailing coefficient of f}  \\
\midrule
 \textbf{Example (2) in Java:} \\
 \tabincell{l}{
 \texttt{$@$SuppressWarnings(STRING)} \\ \texttt{\textcolor[RGB]{59,162,59}{public} \textcolor{blue}{PropagationImp}(\textcolor[RGB]{59,162,59}{Stack}$<$CompositeTransaction$>$lineage, boolean serial, long timeout)\{ } \\ \hspace{2em} \texttt{serial\_ = serial;}  \\ \hspace{2em} \texttt{lineage\_ = (\textcolor[RGB]{59,162,59}{Stack}$<$CompositeTransaction$>$)lineage.clone();} \\ \hspace{2em} \texttt{timeout\_ = timeout;\} }   \\ \textbf{Human-generated}: \textcolor[RGB]{255,0,0}{\textbf{cannot understand}} \\ \textbf{\tool}: create a new instance  \\ \textbf{Ground truth}: create a new instance} \\
\bottomrule
\end{tabular}
}
\end{table*}

To further investigate the quality of auto-generated summaries, we invite participants to summarize the code without access to the reference summary, and then ask the annotators to score the summaries. During manual code summarization, we invite four postgraduate students with more than five years of development experience as well as internship experience in technology companies to participant.
To ease the pressure of annotation, we randomly select 80 code snippets with code lengths fewer than 250 characters. Each participant is asked to write summaries for 20 code snippets, i.e., 10 in Java and 10 in Python. Each of them will be paid 15 USD upon completing the questionnaire.

\begin{table*}[t]
\newcommand{\tabincell}[2]{\begin{tabular}{@{}#1@{}}#2\end{tabular}}
\centering
\caption{Examples illustrating the difference between summaries generated by \tool and written by human.}\label{tab:human_case_study}
\scalebox{0.93}{
\begin{tabular}{l}
\toprule
 \textbf{Example (1) in Python:} \\
 \tabincell{l}{
 \texttt{\textcolor[RGB]{59,162,59}{\textbf{def}} \textcolor{blue}{Thing2Literal}(o, d):   } \\ \hspace{2em} \texttt{return string\_literal(o, d)} \\  \textbf{Human-generated}: return the literalness of a string  \\ \textbf{\tool}: convert something into a string representation  \\ \textbf{Ground truth}: convert something into a sql string literal }\\
\midrule
 \textbf{Example (2) in Python:} \\
 \tabincell{l}{
 \texttt{\textcolor[RGB]{59,162,59}{\textbf{def}} \textcolor{blue}{print\_bucket\_acl\_for\_user}(bucket\_name, user\_email):   } \\ \hspace{2em} \texttt{storage\_client = storage.Client()} \\  \hspace{2em}  \texttt{bucket = storage\_client.bucket(bucket\_name) } \\  \hspace{2em} \texttt{bucket.acl.reload()} \\  \hspace{2em} \texttt{roles = bucket.acl.user(user\_email).get\_roles()} \\  \hspace{2em} \texttt{\textcolor[RGB]{59,162,59}{\textbf{print}} roles}\\  \textbf{Human-generated}: print the bucket acl for user \\ \textbf{\tool}: prints out a buckets access control list for a given user  \\ \textbf{Ground truth}: prints out a buckets access control list for a given user }\\
\midrule
 \textbf{Example (3) in Java:} \\
 \tabincell{l}{
 \texttt{\textcolor[RGB]{59,162,59}{\textbf{public}} \textcolor{blue}{ActivityResolveInfo}(ResolveInfo resolveInfo)\{} \\ \hspace{2em} \texttt{  this.resolveInfo = resolveInfo;\} }\\ \textbf{Human-generated}: assign a value to resolveinfo attribute  \\ 
 \textbf{\tool}: creates a new activity  \\ 
 \textbf{Ground truth}: creates a new instance }\\
\midrule
 \textbf{Example (4) in Java:} \\
 \tabincell{l}{
 \texttt{\textcolor[RGB]{59,162,59}{\textbf{public}} \textcolor[RGB]{59,162,59}{\textbf{static}} Date \textcolor{blue}{parseText}(String dateStr)\{   } \\ \hspace{2em} \texttt{\textcolor[RGB]{59,162,59}{\textbf{try}} \{\textcolor[RGB]{59,162,59}{\textbf{return}} mSimpleTextFormat.parse(dateStr);\} } \\ \hspace{2em} \texttt{\textcolor[RGB]{59,162,59}{\textbf{catch}}(ParseException e)\{ } \\ \hspace{4em} \texttt{e.printStackTrace();} \\ \hspace{4em} \texttt{\textcolor[RGB]{59,162,59}{\textbf{throw new}} RuntimeException(STRING);\}\}} \\ \textbf{Human-generated}: parse the dateStr as a date instance \\ \textbf{\tool}: parse string to datetime  \\ \textbf{Ground truth}: parse string to datetime}\\
\bottomrule
\end{tabular}
}
\end{table*}

\begin{table}[t]
\centering
\aboverulesep=0ex
\belowrulesep=0ex
\caption{Human evaluation on summaries generated by \tool and human-written summaries. }\label{tab:human_written_study}
\scalebox{0.99}{
\begin{tabular}{l|c|cc}
\toprule
Dataset & Metrics & Human-written & \tool \\
\midrule
\multirow{3}{*}{\textbf{\ Java}} & Adequacy &  \textbf{4.38} &3.62\\
& Conciseness  & \textbf{4.82} & 4.48\\
& Fluency  & \textbf{4.94} & 4.58\\
\midrule
\multirow{3}{*}{\textbf{\ Python}} & Adequacy &  \textbf{4.39} & 3.24\\
& Conciseness & \textbf{4.86} & 4.23\\
& Fluency & \textbf{4.95} & 4.34\\
\bottomrule
\end{tabular}
}
\end{table}

We totally receive annotations of 78/80 code snippets. For the remaining two code snippets, as shown in Table~\ref{tab:human_case_study_fail}, the functionalities are hard to be understood by the annotators without corresponding prior knowledge. For the example in Table~\ref{tab:human_case_study_fail} (1), the unclear meanings of ``\textit{K.zero}'' and ``\textit{f[(-1)]}'' hinder the program comprehension. We measure the quality of summaries generated by \tool and humans using the same metrics introduced in Section 4.2, respectively. The results are shown in Table~\ref{tab:results_rating_write}. From the table, we find that compared with human-generated summaries, auto-generated summaries are much more similar to the reference summaries. For example, the BLEU-4 scores of the auto-generated summaries are 77.06 and 75.09 on Java and Python, respectively, while the human-generated summaries are only 19.04 and 23.63 on Java and Python, respectively.
To analyze the reason of the large difference between human-generated summaries and reference summaries, we manually check all the annotated data, and summarize two main reasons:

\begin{itemize}
\item \textbf{Lack of contextual knowledge.} Some code snippets use external APIs or inner elements of a class, and the details of the APIs and elements cannot be accessed. So the annotators can only infer the functions of code snippets based on the function/variable names, resulting in poorly-written summaries. For example, as shown in Table~\ref{tab:human_case_study} (1), since the detail of the external API ``\textit{string\_literal}'' is unknown, humans can only guess its meaning from the name. In Table~\ref{tab:human_case_study} (2), ``\textit{acl}'' is an inner element of the class ``\textit{bucket}'', but the definition is lacking, which makes it hard for humans 
to comprehend the function. Our model has been provided with the knowledge that ``\textit{acl}'' stands for ``\textit{access control list}'' during training, so it can output a more accurate summary.
\item \textbf{Limitation of the evaluation metrics.} As shown in Table~\ref{tab:human_case_study} (3) and (4), although the summaries generated by humans can accurately reflect the functions of the code snippets, they are significantly 
different from the reference summaries, leading to low metric scores. For the example in Table~\ref{tab:human_case_study} (4), both the human-generated summary and reference summary explain the meaning of the code snippet well. However, under the existing metrics based on n-gram matching, the metric scores between them are very low since they have only one overlapping word ``\textit{parse}''.
\end{itemize}

We then qualitatively inspect the quality of human-generated summaries and auto-generated summaries. We invite another three annotators, who have not joined the manual code summarization part, for the inspection. The results are illustrated in Table~\ref{tab:human_written_study}. The Cohen's kappa scores of the annotation results are 0.69 and 0.71 on Java and Python, respectively, indicating a substantial inter-rater reliability on both datasets. As shown in Table~\ref{tab:human_written_study}, the quality of human-generated summaries is better than that of the auto-generated summaries with respect to all the metrics. Specifically, the conciseness and fluency scores of Human-written summary are nearly 5 on both datasets. Moreover, 
the adequacy scores of human-written summaries 
outperform the auto-generated summaries by 21.0\% and 35.5\% on Java and Python datasets, respectively. The results further explain the huge difference between human-generated summaries and reference summaries under automatic evaluation, as shown in Table~\ref{tab:results_rating_write}, reflecting the limitation of automatic metrics.
\section{Discussion}\label{sec:discussion}
In this section, we mainly discuss the key properties of the proposed \tool, the impact of duplicate data in the benchmark dataset on the model performance, and the limitations of our study.

\begin{table*}[t]
\newcommand{\tabincell}[2]{\begin{tabular}{@{}#1@{}}#2\end{tabular}}
\centering
\caption{Examples illustrating summaries generated by different approaches given the code snippets. The following examples are only from the test set and do not exist in the training set.
}\label{tab:case_study}
\scalebox{0.9}{
\begin{tabular}{l}
\toprule
 \textbf{Example (1) in Java:} \\

 \tabincell{l}{
 \texttt{\textcolor[RGB]{59,162,59}{\textbf{public static}} boolean \textcolor{blue}{isFile}(String path)\{  } \\ \hspace{2em}
 \texttt{File f = \textcolor[RGB]{59,162,59}{\textbf{new}} File(path); } \\  \hspace{2em}  
 \texttt{\textcolor[RGB]{59,162,59}{\textbf{return}} f.isFile();  }\\ 
 \texttt{\}}\\  
 \textbf{Vanilla Transformer}: checks if the given path is a file object , is a directory it can be read . no distinction is \\considered exceptions \\ \textbf{CodeTransformer}: checks if the given file is a file\\ \textbf{NeuralCodeSum}: checks if is file exist \\ \textbf{GREAT}: checks if the given path is a file\\  \textbf{\tool}: checks if the given path is a file \\ \textbf{Ground truth}: checks if the given path is a file}  \\
\midrule
 \textbf{Example (2) in Python:} \\
 \tabincell{l}{
 \texttt{\textcolor[RGB]{59,162,59}{\textbf{def}} \textcolor{blue}{print\_bucket\_acl\_for\_user}(bucket\_name, user\_email):   } \\ \hspace{2em} \texttt{storage\_client = storage.Client()} \\  \hspace{2em}  \texttt{bucket = storage\_client.bucket(bucket\_name) } \\  \hspace{2em} \texttt{bucket.acl.reload()} \\  \hspace{2em} \texttt{roles = bucket.acl.user(user\_email).get\_roles()} \\  \hspace{2em} \texttt{\textcolor[RGB]{59,162,59}{\textbf{print}} roles}\\  \textbf{Vanilla Transformer}: removes a user from the access control list \\  \textbf{CodeTransformer}: sets a the user from to to .\\ \textbf{NeuralCodeSum}: prints out a user access control list \\ \textbf{GREAT}: prints out a user access control list \\ \textbf{\tool}: prints out a buckets access control list for a given user  \\ \textbf{Ground truth}: prints out a buckets access control list for a given user }\\
\midrule
 \textbf{Example (3) in Python:} \\
 \tabincell{l}{
 \texttt{\textcolor[RGB]{59,162,59}{\textbf{def}} \textcolor{blue}{token\_urlsafe}(nbytes=None):  } \\ \hspace{2em} 
 \texttt{tok = token\_bytes(nbytes)} \\  \hspace{2em}  
 \texttt{\textcolor[RGB]{59,162,59}{\textbf{return}} base64.urlsafe\_b64encode(tok).rstrip(\textcolor[RGB]{170,0,34}{`='}).decode(\textcolor[RGB]{170,0,34}{`ascii'}) }\\  \textbf{Vanilla Transformer}: generate a token \\ \textbf{CodeTransformer}: decodes a unicode string string string if \\ \textbf{NeuralCodeSum}: construct a random text string . \\ \textbf{GREAT}: generates a token identifier \\  \textbf{\tool}: return a random url-safe string . \\ \textbf{Ground truth}: return a random url-safe text string .}  \\
\bottomrule
\end{tabular}
}
\end{table*}

\subsection{Why Does Our Model Work?}
\begin{figure*}[t]
\centering
\includegraphics[width=1.0\textwidth]{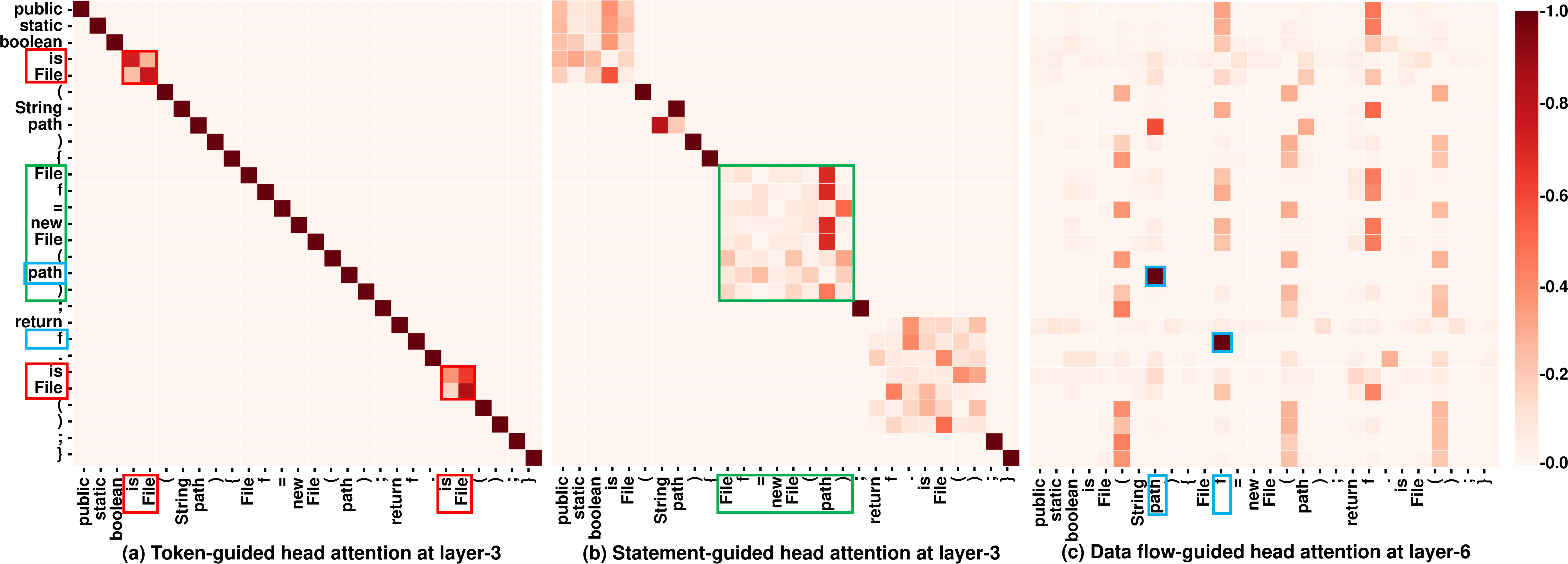}
\caption{Heatmap visualization of self-attention scores of the three types of heads in the encoder for the first case in Table~\ref{tab:case_study}. The rectangles with red edge, green edge, and blue edge indicates the tokens belonging to the same original token, the same statement, or containing data flow relation, respectively.}
\label{fig:atten_vis}
\end{figure*}

\begin{figure*}[t]
\centering
\includegraphics[width=1.0\textwidth]{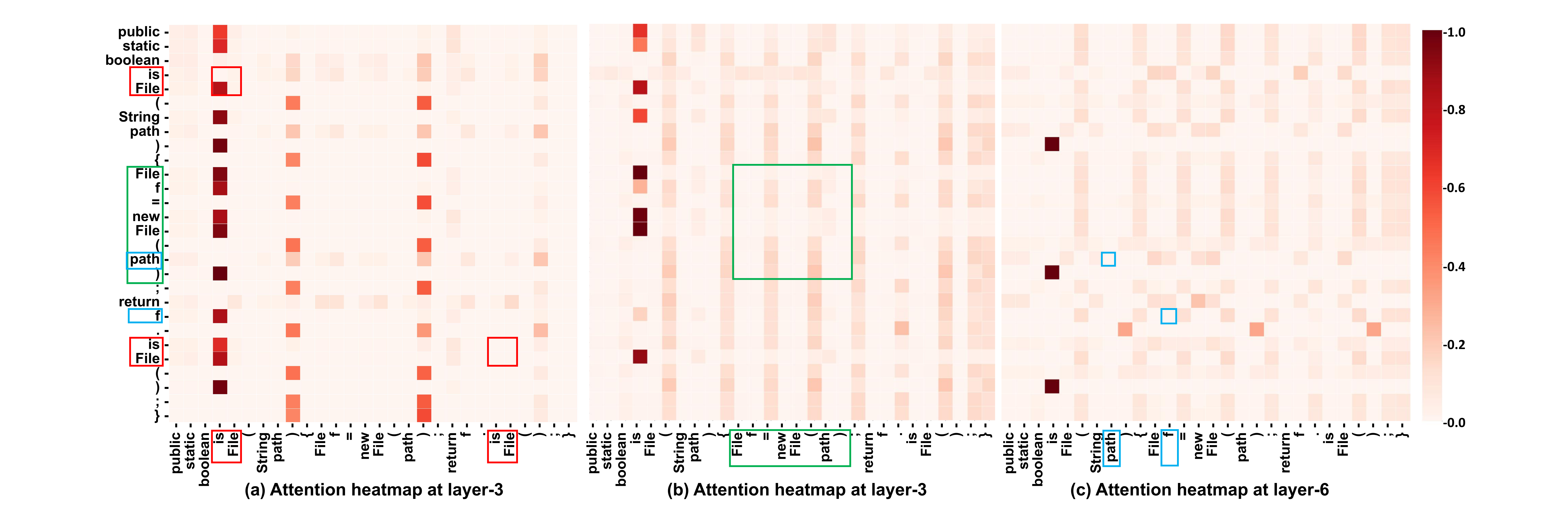}
\caption{Heatmap visualization of self-attention scores of NeuralCodeSum. The rectangles with red edge, green edge, and blue edge indicates the tokens belonging to the same original token, the same statement, or containing data flow relation, respectively.}
\label{fig:ncs_atten_vis}
\end{figure*}

We further conduct an analysis to gain insights into the proposed \tool in generating high-quality code summaries. Through qualitative analysis, we have identified two properties of \tool that may explain its effectiveness in the task.

\textbf{Observation 1: \tool can better capture the semantic relations among tokens.} From Example (1) shown in Table~\ref{tab:case_study}, we can observe that \tool produces the summaries most similar to the ground truth among all the approaches, while the CodeTransformer gives the worst result. We then visualize the heatmap of the self-attention scores of the three types of heads in Figure~\ref{fig:atten_vis} for a further analysis. As can be seen in Figure~\ref{fig:atten_vis} (a) and (b), \tool can focus on local relations among code tokens through its token-guided self-attention and statement-guided self-attention. For example, \tool can learn that the two tokens ``\textit{is}'' and ``\textit{File}'' possess a strong relation, according to Figure~\ref{fig:atten_vis}~(a). As depicted in Figure~\ref{fig:atten_vis} (b), we can find that \tool captures that the token ``\textit{path}'' is strongly related to the corresponding statement, which may be the reason the token ``\textit{path}'' appears in the summary. Figure~\ref{fig:atten_vis} (c) shows that the data flow-guided head attention focuses more on the global information, and can capture the strong relation between the tokens ``\textit{path}'' and ``\textit{f}''. Based on the analysis of the example (1), we {speculate} 
that the model can well capture the token relations locally and globally for code summary generation. As for the heatmap of other baseline models like NeuralCodeSum, we can see that they are very different with the heatmap of SG-Trans. As shown in Figure~\ref{fig:ncs_atten_vis}, NeuralCodeSum does not capture the token, statement, and data flow information in any layer while SG-Trans can pay more attention to the token pairs with syntactic or semantic relations. A similar conclusion can be drawn from Example (2) in Table~\ref{tab:case_study}. All the approaches successfully comprehend that the token ``\textit{acl}'' indicates ``\textit{access control list}''. However, the vanilla Transformer fails to capture the semantic relations between ``\textit{print}'' and ``\textit{acl}'', and NeuralCodeSum misunderstands the relations between ``\textit{user}'' and ``\textit{acl}''. Instead, \tool accurately predicates both relations through the local self-attention and global self-attention.

\textbf{Observation 2: Structural information-guided self-attention can facilitate the copy mechanism to copy important tokens.} In Example (3) in Table~\ref{tab:case_study}, \tool successfully identified the important token ``\textit{urlsafe}'' in the given code while generating the summary. But both vanilla Transformer and NeuralCodeSum ignored the token and output less accurate summaries. The reason that the important token is successfully copied by \tool may be attributed to the structural information-guided self-attention which helps focus on the source tokens more accurately.

\subsection{Duplicate Data in the Java Dataset}
During our experimentation, we find that there are duplicate data in the Java dataset, which may adversely affect the model performance~\citep{DBLP:conf/oopsla/Allamanis19}. As for the Python dataset, there is no duplication across different training, validation and test set. As shown in Table~\ref{tab:duplication_details}, there are 23.3\% and 23.7\% duplicate data in the validation set and the test set, respectively. To evaluate the impact of the data duplication on the proposed model, we remove the duplicate data across the training, validation, and test sets. We choose the two strongest baselines, NeuralCodeSum and GREAT, for comparison. The results after deduplication are shown in Table~\ref{tab:duplication}. As can be seen, all models present a dramatic decrease on the de-duplicated dataset. Nevertheless, the proposed \tool still outperforms GREAT on the BLEU-4, ROUGE-L and METEOR metrics, i.e., by 3.3\%, 2.7\% and 3.3\%, respectively.


\begin{table}
\centering
\caption{Duplicate data in the Java dataset.}\label{tab:duplication_details}
\aboverulesep=0ex
\belowrulesep=0ex
\scalebox{1.0}{
\begin{tabular}{l|cc}
\toprule
 & \textbf{Validation Set} & \textbf{Test set}\\
\midrule
{ Total data} & 8,714 & 8,714 \\
{ Duplicate data } &2,028 (23.3\%) & 2,059 (23.6\%)\\
\bottomrule
\end{tabular}
}
\end{table}

\begin{table}
\centering
\caption{Comparison results on the de-duplicated Java dataset. Data listed within brackets are computed drop rates compared with the results on original Java dataset.}\label{tab:duplication}
\aboverulesep=0ex
\belowrulesep=0ex
\scalebox{1.0}{
\begin{tabular}{l|ccc}
\toprule
{\textbf{Approach}} & BLEU-4 & ROUGE-L & METEOR\\
\midrule
{ NeuralCodeSum} & 29.37 ($\downarrow$34.95\%) & 41.62 ($\downarrow$24.11\%) & 19.98 ($\downarrow$27.24\%)\\
{ GREAT} & 29.49 ($\downarrow$34.52\%) & 41.84 ($\downarrow$23.12\%) & 20.15 ($\downarrow$25.79\%)\\
{ \tool } &\textbf{30.46} ($\downarrow$33.74\%) & \textbf{42.97} ($\downarrow$23.07\%) & \textbf{20.82} ($\downarrow$25.32\%)\\
\bottomrule
\end{tabular}
}
\end{table}

\subsection{Analysis of the Hierarchical Structure-Variant Attention Mechanism}
\begin{figure*}[t]
    \centering
    \begin{subfigure}[b]{0.45\textwidth}
        \includegraphics[width=\textwidth]{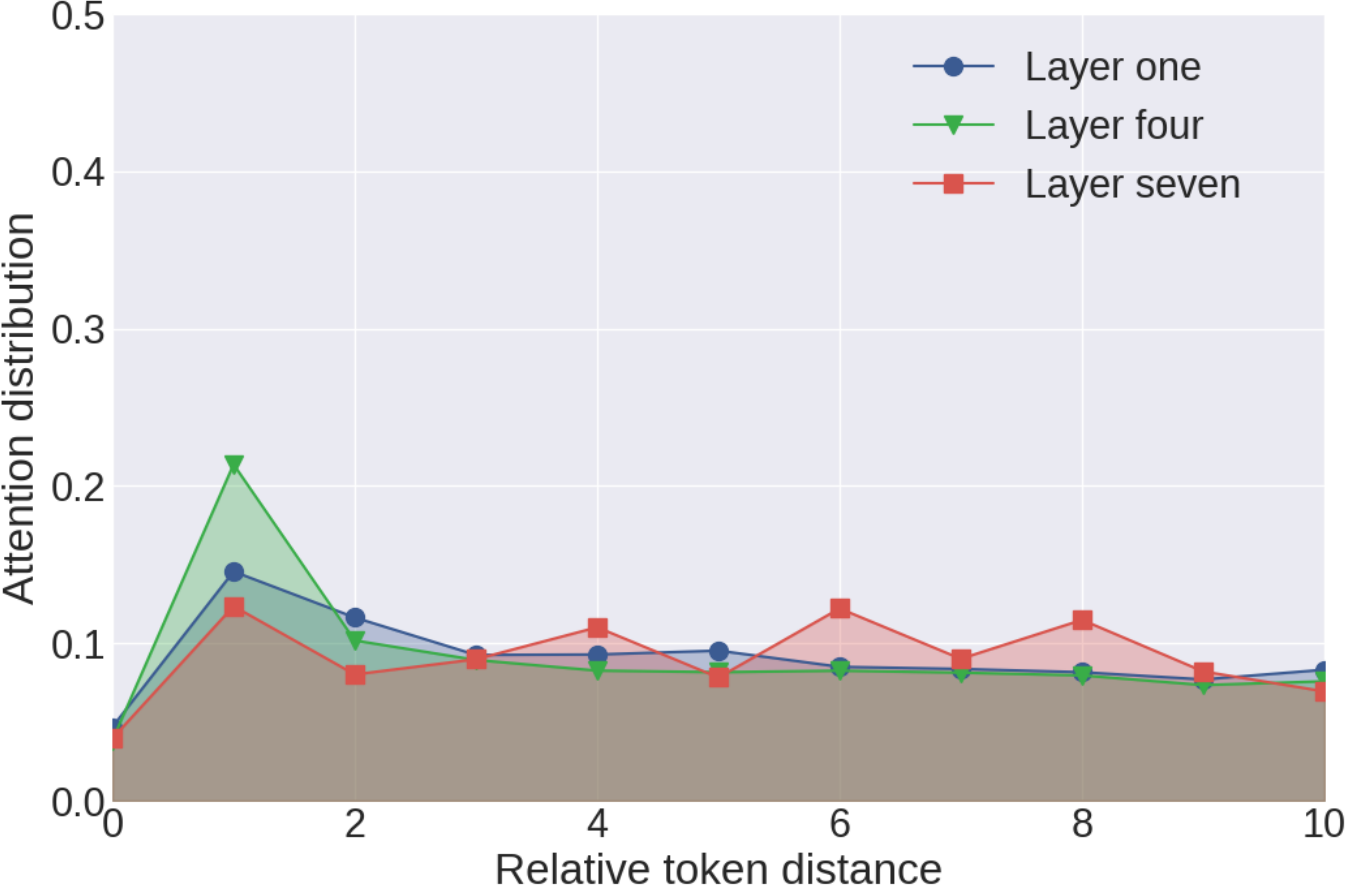}
        \caption{GREAT.}
        \label{tab:adgreat}
      \end{subfigure}
      \hfill
      \begin{subfigure}[b]{0.45\textwidth}
        \includegraphics[width=\textwidth]{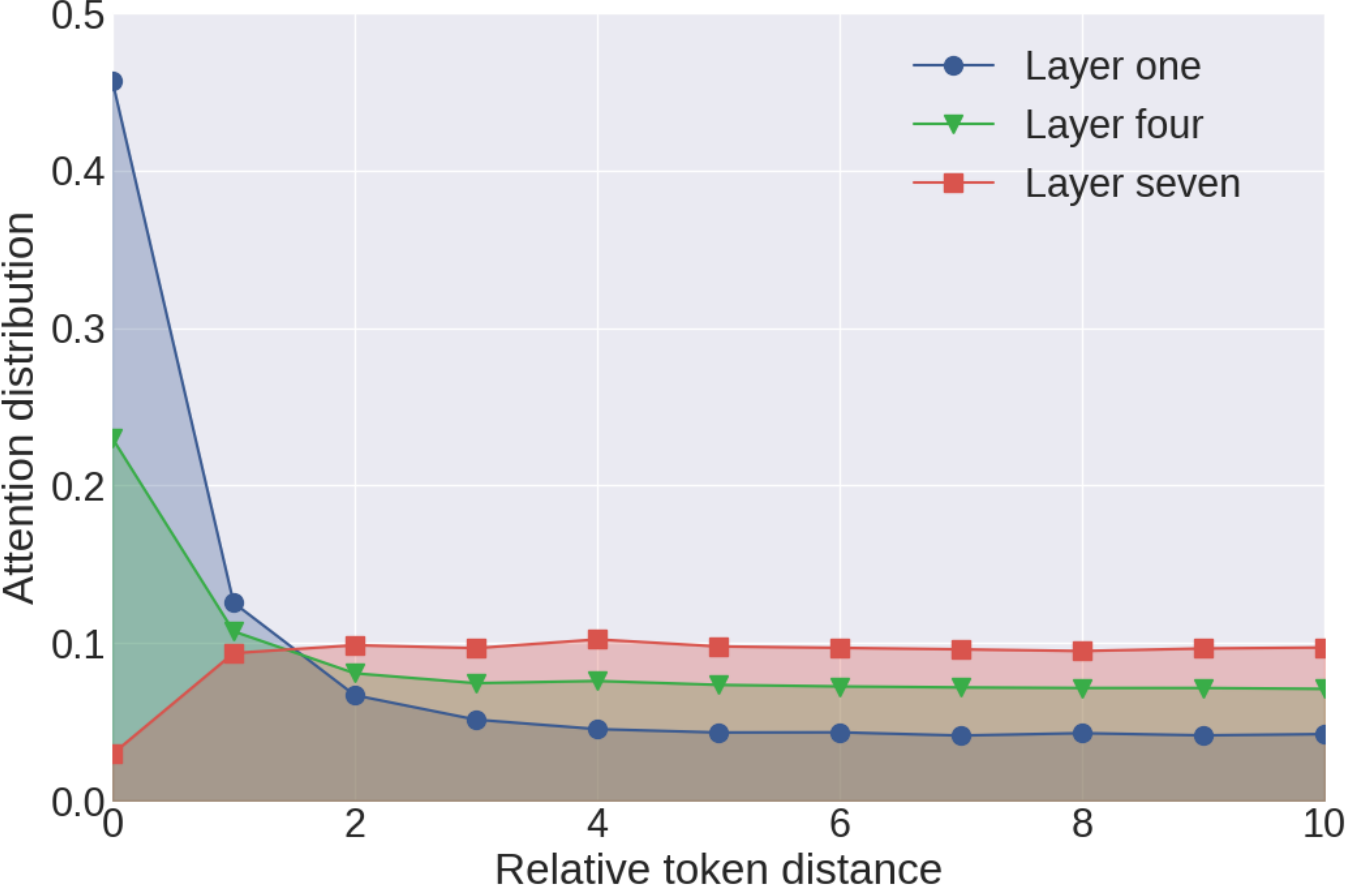}
        \caption{\tool.}
        \label{tab:adsgtrans}
      \end{subfigure}
    \caption{Attention distributions regarding relative token distance for (a) GREAT  and (b) \tool. The horizontal axis represents the relative distance to the current token, and the vertical axis denotes the normalized attention scores along with
    relative distances in one layer.}
	\label{fig:att_dis}
\end{figure*}

The hierarchical structure-variant attention mechanism in \tool aims at rendering the model focus on local structures at shallow layers and global structure at deep layers. In the section, we analyze whether the mechanism can assist \tool learning the hierarchical information. We visualize the distributions of attention scores corresponding to the relative token distances for the shallow layer -- Layer one, middle layer -- Layer four, and one deep layer -- Layer seven, respectively. Specifically, for each relative token distance $\iota$, its attention distribution $Y_\iota$ is computed as Equ. (\ref{equ:1}).

\begin{equation}\label{equ:1}
{ Y }_{\iota} = \frac{\sum_{i=1}^{N} \text{attention}(i,i+\iota)+\text{attention}(i,i-\iota)}{\sum_{j=1}^{S}\sum_{i=1}^{N} \text{attention}(i,i+j)+\text{attention}(i,i-j)}
\end{equation}
where $N$ denotes the number of tokens and $S$ denotes the longest relative distance for analysis ($S=10$ in our analysis). The attention($i,j$) denotes the attention score of token $x_i$ to $x_j$ ($1 \le j \le N$).
The attention scores reflect whether the model focuses on local information or global information.
We choose the relational Transformer GREAT for comparison since it also involves structural information but is not designed hierarchically. The visualization is depicted in Figure~\ref{fig:att_dis}. For GREAT, as shown in Figure~\ref{fig:att_dis} (a), we find that the attention distributions across different layers present similar trends, i.e., they all tend to focus on different token distances uniformly. For \tool, as shown in Figure~\ref{fig:att_dis} (b), we can observe that the three layers pay various attentions to tokens of different relative distances. The shallow layer (Layer one) more focuses on tokens with short relative distances. In the middle layer (Layer four), the attention distribution among different distances is more balanced, which indicates that the model pays increasingly more attention to global tokens with the layer depth being increased. For the deep layer (Layer seven), the attention scores for tokens of long distances are larger than those of short distances, meaning that the model tends to focus on long-range relations in deep layer. The results demonstrate that the hierarchical attention mechanism in \tool is beneficial for the model to capture the hierarchical structural information which cannot be easily learned by the relational Transformer.

\subsection{Difference with Relational Transformers}
The main differences between \tool and the relational Transformers~\citep{DBLP:conf/iclr/HellendoornSSMB20,DBLP:conf/iclr/ZugnerKCLG21} are mainly in two aspects:
 \begin{enumerate}
 \item \textbf{Strategy in 
 incorporating structural information.}
 Compared with GREAT \cite{DBLP:conf/iclr/HellendoornSSMB20} and CodeTransformer \cite{DBLP:conf/iclr/ZugnerKCLG21} which use learnable bias and sinusoidal encoding function to encode the structure information, respectively, 
 \tool incorporates the local and global structure information with different strategies, e.g., introducing the local information with hard mask. The results in Table~\ref{tab:results} and Table~\ref{tab:ablation} show the effectiveness of the structural incorporation strategy in \tool.
 
 \item \textbf{Design of hierarchical structure-variant attention mechanism.} In \tool, a hierarchical attention mechanism is designed to assist model learning the hierarchical information; while the relational Transformers do not involve such design. Both the ablation study in Section 5.2 and the discussion in Section 6.3 demonstrate the benefits of the design.
 \end{enumerate}
 
\subsection{Difference with GraphCodeBERT}

\tool takes data flow information which is similar to GraphCodeBERT~\cite{DBLP:journals/corr/abs-2009-08366}. However, the two methods are different in the following aspects.

 \begin{enumerate}
 
 \item \textbf{Role of data flow.} GraphCodeBERT mainly uses the data flow in two ways: (1) filtering the irrelevant signals in the variable sequence, (2) guiding the two pre-training tasks, including edge prediction and node alignment. Nevertheless, \tool directly uses data flow information to help the attention mechanism better capture the data dependency relations in source code.

 \item \textbf{Incorporation way of data flow.} We integrate the data flow information in a different way as GraphCodeBERT. GraphCodeBERT utilizes a sparse masking mechanism to mask the attention relations between the tokens without data dependency. However, \tool retains the attention relations for the tokens without data dependency, and also highlights the data flow dependency through our designed data flow-guided self-attention.
 
 \item \textbf{Targets of the proposed model.} The targets of the two models are different. GraphCodeBERT is proposed to utilize the inherent structure of code to facilitate the pre-training stage. However, \tool mainly focuses on the task without large amount of source code available and uses the incorporated code structure to alleviate the dependency on the training data.
 \end{enumerate}

\begin{table}[ht]
 \centering
 \aboverulesep=0ex
\belowrulesep=0ex
\caption{Comparison of the cost of different models. The ``\_'' under the preprocessing time of NeuralCodeSum and CodeBERT denotes that the approaches do not need preprocess.  The ``\_'' under the training time of CodeBERT denotes that we do not reproduce the pre-training stage due to the limitation of computing resource.}
\label{tab:cost}
\scalebox{0.99}{
\begin{tabular}{l|ccc}
\toprule
& \textbf{GPU memory usage} & \textbf{Training time} & \textbf{ Preprocessing time }\\ 
\midrule
{\ NeuralCodeSum} & 8729M & 30.4h & -\\
{\ CodeTransformer } & 8573M & 211.5h & 66.1ms\\
{\ \tool } & 8509M & 29.9h & 3.8ms\\
{\ CodeBERT } & 17959M & - & -\\
\bottomrule
\end{tabular}}
\end{table}
 
\subsection{Analysis of the Complexity of \tool}

\tool incorporates structural information based on three types of relations between code tokens. Comparing with the baseline approaches, such as NeuralCodeSum, \tool involves more types of relations, which could lead to an increase in the model complexity and subsequently impacting its applicability to other programming languages. To investigate to what extent \tool introduces extra complexity, we conduct analysis of the cost of \tool.

Specifically, we first compare the cost of \tool with Transformer-based baselines including NeuralCodeSum and CodeTransformer, and a pre-training model CodeBERT, in terms of the GPU memory usage, training time cost and preprocessing time cost. The comparison is implemented on the same server with a single Tesla V100 GPU by training on the Java dataset with 32 batch size. The results are shown in Table~\ref{tab:cost}. As can be seen, the GPU memory usage and training time cost of \tool are the lowest among all the approaches. Since \tool does not involve the relative positive embedding, both its GPU memory usage and training time cost are even lower than NeuralCodeSum. Table~\ref{tab:cost} also shows that CodeBERT requires the highest memory usage, restricting its application to low-resource devices. With respect to the preprocessing time cost for one sample, since \tool does not need calculate the complex features used in CodeTransformer such as shortest path length and personalized PageRank, it only takes about 3.8ms which is significantly faster than CodeTransformer. The results indicate that the code structure properties used in \tool do not bring larger cost than the baselines.


For the application of \tool to other programming languages, the main barrier lies in the data flow extraction procedure. In this work, we follow the main idea of Wang et al.'s work~\cite{DBLP:conf/wcre/WangLM0J20} and only consider the common data flow information which is generally similar for different programming languages. The common data flow information includes sequential data flow relations and three types of non-sequential data flow relations such as ``\textit{if}'' statements, ``\textit{for}'' and ``\textit{while}'' loops. The sequential data flow relations can be easily extracted by identifying the variables for any programming language. For the non-sequential data flow relations, the extraction procedure of different programming languages is also similar. Because the AST parser tree-sitter\footnote{https://github.com/tree-sitter/tree-sitter} can parse the data flow relations of different languages into almost the same tree structure. Thus, it is convenient to extend \tool to other popular languages.

\subsection{Comparison with CodeBERT}

Many pre-training models \citep{DBLP:conf/emnlp/FengGTDFGS0LJZ20,DBLP:journals/corr/abs-2009-08366} have been proposed recently, which can be adopted for source code summarization. So we also compare the performance of \tool with the most typical pre-training model CodeBERT. We also train \tool under the same model size as CodeBERT and denote it as SG-Trans\textsubscript{large} for further comparison.

As shown in Table~\ref{tab:results_bert}, without fine tuning, CodeBERT shows the worst performance among all the approaches. After enough fine tuning, CodeBERT improves a lot and even outperforms \tool on some metrics, e.g., the METEOR score on the Java dataset. However, it should be noted that the encoder layer number and hidden size of CodeBERT are much larger than \tool. Specifically, the numbers of encoder layers and hidden size of CodeBERT are 10 and 768, respectively; while \tool only has 8 encoder layers and hidden size as 512. For fair comparison, we also train \tool with the same model settings as CodeBERT, denoted as SG-Trans\textsubscript{large}. As shown in Table~\ref{tab:results_bert}, SG-Trans\textsubscript{large} obtains the best performance almost all the metrics. Specifically, on the Java dataset, SG-Trans\textsubscript{large} outperforms CodeBERT+fine-tune by 4.2\% and 1.3\% in terms of BLEU-4 and ROUGE-L, respectively. And on the Python dataset, SG-Trans\textsubscript{large} is only a little bit lower than CodeBERT+fine-tune on ROUGE-L but obviously outperforms CodeBERT+fine-tune by 4.7\% in terms of the BLEU-4 score. The results demonstrate that \tool is more effective than CodeBERT even with accessing to limited data.

\begin{table}[t]
\centering
\caption{Comparison results with CodeBERT. The ``CodeBERT'' represents the the CodeBERT approach without fine tuning. The ``SG-Trans\textsubscript{large}'' represents \tool with the same model settings as CodeBERT, i.e., 10 encoder layers and hidden size as 768.
}\label{tab:results_bert}
\aboverulesep=0ex
\belowrulesep=0ex
\scalebox{1.0}{
\begin{tabular}{l|ccc|ccc}
\toprule
\multirow{2}{*}{\textbf{Approach}} & \multicolumn{3}{c|}{\textbf{Java}} & \multicolumn{3}{c}{\textbf{Python}} \\
\cmidrule{2-7} 
& BLEU-4 & METEOR & ROUGE-L &  BLEU-4 & METEOR & ROUGE-L\\
\midrule
{ CodeBERT\cite{DBLP:conf/emnlp/FengGTDFGS0LJZ20}  } &14.93   & 9.23 & 30.43 &16.70  & 9.68 & 30.31\\
{ CodeBERT+fine-tune\cite{DBLP:conf/emnlp/FengGTDFGS0LJZ20}  } &44.40  & 28.33 &  55.56 & 32.04  & 20.77 & \textbf{47.45}\\
\midrule
{ \tool } &45.89 & 27.85 & 55.79 & 33.04 & 20.52 & 47.01  \\
{ SG-Trans\textsubscript{large} } &\textbf{46.27} & \textbf{28.37} & \textbf{56.30} & \textbf{33.53} & \textbf{20.87} & 47.42  \\
\bottomrule
\end{tabular}
}
\end{table}
\subsection{Threats to Validity}\label{subsec:threat}
There are three main threats to the validity of our study.

\begin{enumerate}
   \item The generalizability of our results. We use two public large datasets, which include 87,136 Java and 92,545 Python code-summary pairs, following the prior research~\cite{DBLP:conf/nips/WeiL0FJ19,DBLP:conf/acl/AhmadCRC20,DBLP:conf/icse/ZhangW00020}. The limited types of programming languages may hinder the scalability of the proposed \tool. In our future work, we will experiment with more large-scale datasets with different programming languages.\par
   
   \item More code structure information could be considered. \tool only takes the token-level and statement-level syntactic structure and the data flow structure into consideration, since it has been previously demonstrated that the data flow information is more effective than AST and CFG during code representation learning~\cite{DBLP:journals/corr/abs-2009-08366}. Nevertheless, other code structural properties such as AST and CFG, could be potentially useful for boosting the model performance. 
In the future, we will explore the use of more structural properties in \tool.
   
   \item Biases in human evaluation. We invited 10 participants to evaluate the quality of 200 randomly selected code-summary pairs. The results of human annotations can be impacted by the participants' programming experience and their understanding of the evaluation metrics. To mitigate the bias of human evaluation, we ensure that the 10 participants are all software developers with at least four years programming experience, and each code-summary pair was evaluated by 3 participants. Summaries generated by different approaches were also randomly shuffled in order to eliminate the order bias. In the future, we will expand the pool of human participants and will also increase the size of the evaluation set.
\end{enumerate}

\section{Related Work}\label{sec:literature}

In this section, we elaborate on two threads of related work, including source code summarization, code representation learning.

\subsection{Source Code Summarization}

There have been extensive research in source code summarization, including template-based approaches \citep{DBLP:conf/kbse/SridharaHMPV10,DBLP:conf/iwpc/MorenoASMPV13,DBLP:journals/tse/McBurneyM16}, information-retrieval-based approaches \citep{DBLP:conf/icse/HaiducAM10,DBLP:conf/acl/Movshovitz-AttiasC13,DBLP:conf/wcre/WongLT15} and deep-learning-based-approaches \citep{DBLP:conf/acl/IyerKCZ16,DBLP:conf/iwpc/HuLXLJ18,DBLP:conf/iclr/AlonBLY19}. Among these categories, deep-learning-based methods have achieved the greatest success and become the most popular in recent years, which specifically formulate the code summarization task as a neural machine translation (NMT) problem and adopt state-of-the-art NMT frameworks to improve the performance. In this section, we focus on deep-learning-based methods and introduce them by their category. We also list an overview of the category of related works in Table~\ref{tab:literature_2}.

\textbf{RNN-based models:}
Iyer et al.~\citep{DBLP:conf/acl/IyerKCZ16} first propose CODE-NN, a Long Short Term Memory (LSTM) network with attention to generate code summaries from code snippets. In order to achieve more accurate code modeling, later researchers then introduce more structural and syntactic information to the deep learning models. Hu et al.~\citep{DBLP:conf/iwpc/HuLXLJ18} propose a structure-based traversal(SBT) method to traverse AST and processing the AST nodes into sequences that can be fed into a RNN encoder. Another work \citep{DBLP:conf/ijcai/HuLXLLJ18} hold the view that code API carries vital information about the functionality of the source code and incorporate the API knowledge by adding an API Sequences Encoder.

\textbf{Tree/GNN-based models:}
To leverage the tree structures of AST, a multi-way Tree-LSTM \citep{DBLP:conf/ijcnn/ShidoKYMM19} is proposed to directly model the code structures. For more fine-grained intra-code relationship exploitation, many works also incorporate code-related graphs and GNN to boost performance. Fernandes et al.~\citep{DBLP:conf/iclr/FernandesAB19} build a graph from source code and extract nodes feature with gated graph neural network while LeClair et al.~\citep{DBLP:conf/iwpc/LeClairHWM20} directly obtain code representation from AST with Convolutional Graph Neural Networks. 
To help model capture more global interactions among nodes, a recent work ~\citep{DBLP:conf/iclr/LiuCXS021} propose a hybrid GNN which fuse the information from static and dynamic graphs via hybrid message passing.

\textbf{Transformer-based models:}
With the rise of Transformer in NMT task domain, Ahmad et al.~\citep{DBLP:conf/acl/AhmadCRC20} equip transformer with copy attention and relative position embedding for better mapping the source code to their corresponding natural language summaries. To leverage the code structure information into Transformer, Hellendoorn et al.~\citep{DBLP:conf/iclr/HellendoornSSMB20} propose GREAT which encode structural information into self attention with adding a learnable edge type related bias. Another work proposed by  Z{\"{u}}gner et al.~\citep{DBLP:conf/iclr/ZugnerKCLG21} focuses on multilingual code summarization and proposes to build upon language-agnostic features such as source code and AST-based features. Wu et al.~\citep{DBLP:conf/acl/WuZZ21} propose the Structure-induced Self-Attention to incorporate multi-view structure information into self-attention mechanism. To capture both the sequential and structural information, a recent work \citep{DBLP:conf/acl/ChoiBNL21} applies graph convolution to obtain structurally-encoded node representations and passes sequences of the graph-convolutioned AST nodes into Transformer. Another recent work~\citep{DBLP:journals/corr/abs-2202-06521} proposes to utilize AST relative positions to augment the structural correlations between code tokens.



\textbf{Information retrieval auxiliary methods:}
The information retrieval auxiliary methods utilize information retrieval and large-scale code repositories to help model generate high-quality summary. Zhang et al.~\citep{DBLP:conf/icse/ZhangW00020} propose to improve code summarization with the help of two retrieved code using syntactic and semantic similarity. To help model generate more important but low-frequency tokens, Wei et al.~\citep{DBLP:conf/kbse/Wei19} further use the existing comments of retrieved code snippets as exemplars to guide the summarization.


\textbf{Multi-task learning strategy:}
Some research try to exploit commonalities and differences across code-related tasks to further improve the code summarization. Wei et al.~\citep{DBLP:conf/nips/WeiL0FJ19} use a dual learning framework to apply the relations between code summarization and code generation and improve the performance of both tasks. A recent work \citep{DBLP:conf/iwpc/00030SZ21} use method name prediction as an auxiliary task and design a multi-task learning approach to improve code summarization performance.

Compared with existing work, our proposed model focuses on improving the Transformer architecture for source code to make it better incorporate both local and global code structure. Other improvement methods like information retrieval and multi-task learning which are orthogonal to our work are not the research target of this paper.

\begin{table}
\centering
\caption{An overview of existing works related to code summarization. The ``IR'' and ``MT'' denotes information retrieval auxiliary methods and multi-task learning strategy respectively.}\label{tab:literature_2}
\aboverulesep=0ex
\belowrulesep=0ex
\scalebox{0.95}{
\begin{tabular}{l|c|c|c|c|c|c}
\toprule
{\textbf{Approach}} & \textbf{Input code property} & \textbf{RNN} & \textbf{Tree/GNN} & \textbf{Transformer} & \textbf{IR} & \textbf{MT}\\
\midrule
CODE-NN \citep{DBLP:conf/acl/IyerKCZ16} & -- & $\checkmark$  &  &  &  & \\
DeepCom \citep{DBLP:conf/iwpc/HuLXLJ18} & AST & $\checkmark$  &  &  &  & \\
Code2Seq \citep{DBLP:conf/iclr/AlonBLY19} & AST & $\checkmark$  &  &  &  & \\
API+Code \citep{DBLP:conf/ijcai/HuLXLLJ18} & API Information & $\checkmark$  &  &  &  & \\
Dual Model \citep{DBLP:conf/nips/WeiL0FJ19} & -- & $\checkmark$  & & &  & $\checkmark$\\
Rencos \citep{DBLP:conf/icse/ZhangW00020} & -- & $\checkmark$ &  &  &  $\checkmark$ & \\
Re$^2$Com \citep{DBLP:conf/kbse/Wei19} & -- & $\checkmark$ &   &  & $\checkmark$ & \\
Tree-LSTM \citep{DBLP:conf/ijcnn/ShidoKYMM19} & AST &  & $\checkmark$ &  &  & \\
Code+GNN \citep{DBLP:conf/iwpc/LeClairHWM20} & AST &  & $\checkmark$ &  &  & \\
HGNN \citep{DBLP:conf/iclr/LiuCXS021} & CPG &  & $\checkmark$ & & $\checkmark$ & \\
NeuralCodeSum \citep{DBLP:conf/acl/AhmadCRC20} & -- &   &  & $\checkmark$ &  & \\
DMACOS \citep{DBLP:conf/iwpc/00030SZ21} & -- &   &  & $\checkmark$ &  & $\checkmark$\\
GREAT \citep{DBLP:conf/iclr/HellendoornSSMB20} &  Multi relations in code &   &  & $\checkmark$ &  & \\
CodeTransformer \citep{DBLP:conf/iclr/ZugnerKCLG21} & AST &   &  & $\checkmark$ &  & \\
Transformer+GNN \citep{DBLP:conf/acl/ChoiBNL21} & AST &  & $\checkmark$ &  $\checkmark$ &  & \\
\bottomrule
\end{tabular}
}
\end{table}

\subsection{Code Representation Learning}

Learning high-quality code representations is of vital importance for deep-learning-based code summarization. Apart from the above practices for code summarization, there also exist other code representation learning methods that lie in similar task domains such as source code classification, code clone detection, commit message generation, etc. For example, the ASTNN model proposed by Zhang et al.~\citep{DBLP:conf/icse/ZhangWZ0WL19} splits large ASTs into sequences of small statement trees, which are further encoded into vectors as source code representations. This model is further applied on code classification and code clone detection. Alon et al.~\citep{DBLP:conf/iclr/AlonBLY19} present CODE2SEQ that represents the code snippets by sampling certain paths from the ASTs. Gu et al.~\citep{DBLP:journals/nn/GuLGWZXL21} propose to encode statement-level dependency relations through the Program Dependency Graph (PDG). Comparatively, the above research on the model architecture improvement for code representation learning is relevant to us but mainly focuses on other code-related tasks such as code classification, code clone detection, etc.

Recently, inspired by the successes of  pre-training techniques in natural language processing field, Feng et al.~\citep{DBLP:conf/emnlp/FengGTDFGS0LJZ20} and Guo et al.~\citep{DBLP:journals/corr/abs-2009-08366} also apply pre-training models on learning source code and achieve empirical improvements on a variety of tasks. To extend the code representations to characterize programs' functionalities, \citep{DBLP:journals/corr/abs-2007-04973} further enriches the pre-training tasks to learn task-agnostic semantic code representations from textually divergent variants of source programs via contrastive learning. The target of the research about pre-training is greatly different from us, i.e., they mainly focus on learning from large-scale dataset in a self-supervised way. However,
our research concentrates on improving the quality of generated summaries with
limited training data in a supervised way.

\section{Conclusion}\label{sec:con}
In this paper, we present \tool, a Transformer-based architecture with structure-guided self-attention and hierarchical structure-variant attention. \tool can attain better modeling of the code structural information, including local structure in token-level and statement-level, and global structure, i.e., data flow. The evaluation on two popular benchmarks suggests that \tool outperforms competitive baselines and achieves state-of-the-art performance on code summarization. 
For future work, we plan to extend the use of our model to other task domains, and possibly build up more accurate code representations for general usage.

\begin{acks}
This research was supported by National Natural Science Foundation of China under project No. 62002084, Stable support plan for colleges and universities in Shenzhen under project No. GXWD20201230155427003-20200730101839009, and the Research Grants Council of the Hong Kong Special Administrative Region, China (No. CUHK 14210920 of the General Research Fund). This research was also partly funded by the UK Engineering and Physical Sciences Research Council (No. EP/V048597/1, EP/T017112/1).  Yulan He is supported by a Turing AI Fellowship funded by the UK Research and Innovation (No. EP/V020579/1). 
\end{acks}


\bibliographystyle{ACM-Reference-Format}
\bibliography{sample-base}


\end{document}